\documentclass[aps,pre,twocolumn,superscriptaddress,showpacs]{revtex4-2}
\usepackage{graphicx}
\usepackage{array}
\usepackage{amsfonts}
\usepackage{amssymb,amsmath,multirow,rotate}
\usepackage{mathrsfs}
\usepackage{booktabs}
\usepackage{threeparttable}
\usepackage{multirow}
\usepackage{epsfig}
\usepackage{threeparttable}
\usepackage{chngpage}
\usepackage{latexsym}
\usepackage{dcolumn}
\usepackage{graphics,graphicx,bm,fleqn,epic,eepic,float}
\usepackage{verbatim}
\usepackage{subfigure}
\usepackage{color}
\usepackage{xr}
\usepackage{float}
\usepackage{listings} 
\usepackage{placeins} 
\usepackage{tikz}
\usepackage{url}
\usepackage{soul}
\usepackage{natbib}

\usepackage{float}
\makeatletter
\let\newfloat\newfloat@ltx
\makeatother

\definecolor{red}{rgb}{1,0,0}

\definecolor{blue}{rgb}{0,0,1}

\begin{document}


\title{Reconstructing dynamics from sparse observations with no training on target system}

\date{\today}

\author{Zheng-Meng Zhai}
\affiliation{School of Electrical, Computer and Energy Engineering, Arizona State University, Tempe, AZ 85287, USA}

\author{Jun-Yin Huang}
\affiliation{School of Electrical, Computer and Energy Engineering, Arizona State University, Tempe, AZ 85287, USA}

\author{Benjamin D. Stern}
\affiliation{Doctor of Physical Therapy Program, Tufts University School of Medicine, 101 E Washington St Suite 950, Phoenix, AZ 85004, USA}

\author{Ying-Cheng Lai} \email{Ying-Cheng.Lai@asu.edu}
\affiliation{School of Electrical, Computer and Energy Engineering, Arizona State University, Tempe, AZ 85287, USA}
\affiliation{Department of Physics, Arizona State University, Tempe, Arizona 85287, USA}

\begin{abstract}

	In applications, an anticipated situation is where the system of interest has never been encountered before and sparse observations can be made only once. Can the dynamics be faithfully reconstructed from the limited observations without any training data? This problem defies any known traditional methods of nonlinear time-series analysis as well as existing machine-learning methods that typically require extensive data from the target system for training. We address this challenge by developing a hybrid transformer and reservoir-computing machine-learning scheme. The key idea is that, for a complex and nonlinear target system, the training of the transformer can be conducted not using any data from the target system, but with essentially unlimited synthetic data from known chaotic systems. The trained transformer is then tested with the sparse data from the target system. The output of the transformer is further fed into a reservoir computer for predicting the long-term dynamics or the attractor of the target system. The power of the proposed hybrid machine-learning framework is demonstrated using a large number of prototypical nonlinear dynamical systems, with high reconstruction accuracy even when the available data is only 20\% of that required to faithfully represent the dynamical behavior of the underlying system. The framework provides a paradigm of reconstructing complex and nonlinear dynamics in the extreme situation where training data does not exist and the observations are random and sparse.

\end{abstract}

\maketitle

\section{Introduction} \label{sec:intro}

In applications of complex systems, observations are fundamental to tasks such as 
mechanistic understanding, dynamics reconstruction, state prediction, and control. 
When the available data are complete in the sense that the data points are sampled 
according to the Nyquist criterion and no points are missing, it is possible to 
extract the dynamics or even find the equations of the system from data by sparse 
optimization~\cite{wang2011predicting,Lai:2021}. In machine learning, reservoir 
computing has been widely applied to complex and nonlinear dynamical systems for 
tasks such as prediction~\cite{HSRFG:2015,LBMUCJ:2017,PLHGO:2017,LPHGBO:2017,PHGLO:2018,Carroll:2018,NS:2018,ZP:2018,GPG:2019,JL:2019,TYHNKTNNH:2019,FJZWL:2020,ZJQL:2020,KKGGM:2020,CLAC:2020,KFGL:2021a,PCGPO:2021,KLNPB:2021,FKLW:2021,KFGL:2021b,Bollt:2021,GBGB:2021,KWGHL:2023,ZKL:2023,YHBTLS:2024}, control~\cite{ZMKGHL:2023}, signal detection~\cite{ZMKL:2023}, and estimation~\cite{ZMGHL:2024}.
Quite recently, Kolmogorov-Arnold networks (KANs)~\cite{liu2024kan}, typically 
small neural networks, were proposed for discovering the dynamics from data, where 
even symbolic regression is possible in some cases to identify the exact mathematical 
equations and parameters. It has also been demonstrated~\cite{MPBL:2024} that the 
KANs have the power of uncovering the dynamical system in situations where the methods
of sparse optimization fail. In all these applications, an essential requirement is
that the time-series data are complete in the Nyquist sense.     

A challenging but not uncommon situation is where a new system is to be learned and 
eventually controlled based on limited observations. Two significant difficulties 
arise in this case. First, being ``new'' means that the system has not been observed 
before, so no previous data or recordings exist. If one intends to exploit machine 
learning to learn and reconstruct the dynamics of the system from observations, no 
training data are available. Second, the observations may be irregular and sparse: the 
observed data are not collected at some uniform time interval, e.g., as determined
by the Nyquist criterion, but at random times with the total data amount much less
than that from Nyquist sampling. It is also possible that the observations can be 
made only once. The question is, provided with one-time sparse observations or 
time-series data, can the dynamics of the underlying system still be faithfully 
reconstructed?

Limited observations or data occur in various real-world situations. For example, 
ecological data gathered from diverse and dynamic environments inevitably contain 
gaps caused by equipment failure, weather conditions, limited access to remote 
locations, and temporal or financial constraints. Similarly, in medical systems
and human activity tracking, data collection frequently suffers from issues such 
as patient noncompliance, recording errors, loss of followup, and technical 
failures. Wearable devices present additional challenges, including battery 
depletion, user error, signal interference from clothing or environmental factors, 
and inconsistent wear patterns during sleep or specific activities where devices 
may need to be removed. A common feature of 
these scenarios is that the available data are only from random times without any 
discernible patterns. This issue becomes particularly problematic when the data 
is sparse. Being able to reconstruct the dynamics from sparse and random data is 
particularly challenging for nonlinear dynamical systems due to the possibility of 
chaos leading to a sensitive dependence on small errors. For example, large errors 
may arise when predicting the values of the dynamical variables in various intervals 
in which data are missing. However, if training data from the {\em same} target 
system are available, machine learning can be effective for reconstructing the 
dynamics from sparse data~\cite{yeo2019data}. (Additional background on 
machine-learning approaches is provided in Appendix~\ref{appendix_na}.)

It is necessary to define what we mean by ``random and sparse'' data. We consider 
systems whose dynamics occur within certain finite frequency band. For chaotic 
systems with a broad power spectrum, in principle the ``cutoff'' frequency can 
be arbitrarily large, but power contained in a frequency range near and beyond the 
cutoff frequency can often be significantly smaller than that in the low frequency 
domain and thus can be neglected, leading realistically to a finite yet still large
bandwidth. A meaningful Nyquist sampling frequency can then be defined. An 
observational dataset being complete means that the time series are recorded 
at the {\em regular} time interval as determined by the Nyquist frequency with 
no missing points. In this case, the original signal can be faithfully 
reconstructed. Random and sparse data mean that the data points are sampled 
at {\em irregular} time intervals and some portion of the data points as determined 
by the Nyquist frequency are missing at random times. We aim to reconstruct the 
system dynamics from random and sparse observations by developing a machine-learning 
framework to generate continuous time series that meet the Nyquist criteria, i.e.,
time series represented by regularly sampled data points of frequency at or exceeding 
the Nyquist frequency. When the governing 
equations of the underlying system are unknown and/or when historical observations 
of the full dynamical trajectory of the system are not available, the resulting lack 
of any training data makes the reconstruction task extremely challenging. 
Indeed, since the system cannot be fully measured and only irregularly observed 
data points are available, directly inferring the dynamical trajectory from these 
points is infeasible. Furthermore, the extent of the available observed data points 
and the number of data points to be interpolated can be uncertain. 

In this paper, we develop a machine-learning framework to address the problem of 
dynamics reconstruction and prediction from random and sparse observations with no 
training data from the target system. Our key innovation is training a hybrid 
machine-learning framework in a laboratory environment using a variety of synthetic 
dynamical systems other than data from the target system itself, and deploy the 
trained architecture to reconstruct the dynamics of the target system from one-time 
sparse observations. 
More specifically, we exploit the machine-learning framework of transformers with 
training data {\em not from the target system} but from a number of known, synthetic 
systems that show qualitatively similar dynamical behaviors to those of the target 
system, for which complete data are available. The training process can thus be 
regarded as a ``laboratory-calibration'' process during which the transformer learns 
the dynamical rules generating the synthetic but complete data. The so-trained 
transformer is then deployed to the real application with the random and sparse data, 
and is expected to adapt to the unseen data and reconstruct the underlying dynamics. 
To enable long-term prediction of the target system, we exploit reservoir computing 
that has been demonstrated to be particularly suitable for predicting nonlinear dynamics~\cite{HSRFG:2015,LBMUCJ:2017,PLHGO:2017,LPHGBO:2017,PHGLO:2018,Carroll:2018,NS:2018,ZP:2018,GPG:2019,JL:2019,TYHNKTNNH:2019,FJZWL:2020,ZJQL:2020,KKGGM:2020,CLAC:2020,KFGL:2021a,PCGPO:2021,KLNPB:2021,FKLW:2021,KFGL:2021b,Bollt:2021,GBGB:2021,KWGHL:2023,ZKL:2023,YHBTLS:2024} by feeding the output of the transformer into the reservoir computer. This 
combination of transformer and reservoir computing constitutes a hybrid 
machine-learning framework. We demonstrate that it can successfully reconstruct 
the dynamics of approximately three dozen prototypical nonlinear systems, with high 
reconstruction accuracy even when the available data is only 20\% of that required 
to faithfully represent the dynamical behavior of the underlying system according 
to the Nyquist criterion. Our framework provides a paradigm of reconstructing complex 
and nonlinear dynamics in the extreme situation where training data from the target 
system do not exist and the observations or measurements are severely insufficient.

\section{Hybrid machine learning} \label{sec:methods}

\begin{figure*} [ht!]
\centering
\includegraphics[width=\linewidth]{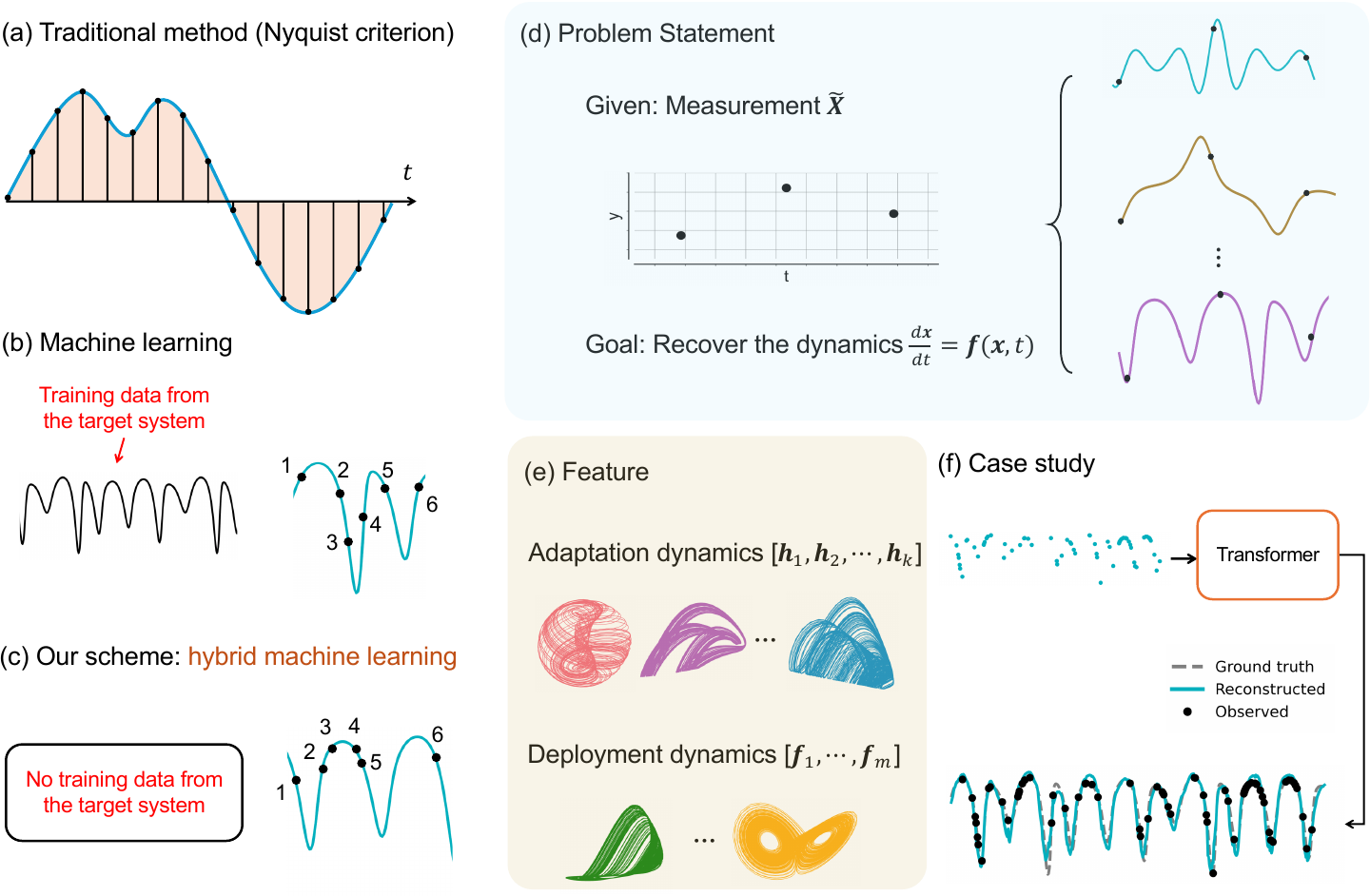} 
\caption{Dynamics reconstruction from random and sparse data. 
(a) The textbook case of a random time series 
sampled at a frequency higher than the Nyquist frequency, where the dynamical data
can be faithfully reconstructed. (b) Training data from the target system (left) 
and a segment of time series of six data points in a time interval containing 
approximately two cycles of oscillation. According to the Nyquist criterion, the 
signal can be faithfully reconstructed with more than 20 uniformly sampled data 
points (see text). When the data points are far fewer than 20 and even worse, they 
are randomly sampled, reconstruction becomes challenging. However, if training data 
from the same target system are available, existing machine-learning methods can 
be used to reconstruct the dynamics from the sparse data~\cite{yeo2019data}. (c) If 
no training data from the target system are available, hybrid machine learning 
proposed here provides a viable solution to reconstructing the dynamics from sparse 
data. (d) Problem statement. Given random and sparse data, the goal is to reconstruct 
the dynamics of the target system governed by $d\mathbf{x}/dt=f(\mathbf{x},t)$. A 
hurdle that needs to be overcome is that, for any given three points, there exist 
infinitely many ways to fit the data, as illustrated on the right side. (e) Training 
of the machine-learning framework is done using complete data from a large number of 
synthetic dynamical systems $[\mathbf{h}_1,\mathbf{h}_2,\cdots,\mathbf{h}_k]$. The
framework is then adapted to reconstruct and predict the dynamics of the target 
systems $[\mathbf{f}_1,\cdots,\mathbf{f}_m]$. (f) An example: in the testing 
(deployment) phase, sparse observations are provided to the trained neural network 
for dynamics reconstruction.} 
\label{fig:main_idea}
\end{figure*}

Figure~\ref{fig:main_idea} highlights the challenge of reconstructing the dynamics
from sparse data without training data. In particular, Fig.~\ref{fig:main_idea}(a)
shows the textbook case of a random time series sampled at a frequency higher than 
the Nyquist frequency, which can be completely reconstructed. To illustrate random
and sparse data in an intuitive setting, we consider a set of six available data 
points from a unit time interval, as shown in Figs.~\ref{fig:main_idea}(b) and 
\ref{fig:main_idea}(c). The time interval contains approximately two periods of 
oscillation, which defines a local frequency denoted as $f_{\rm local} = 2$. As the 
signal is chaotic or random, the cutoff frequency $f_{\rm max}$ in the power 
spectrum can be higher than the frequency represented by the two oscillation cycles 
as shown. As a concrete example, we assume $f_{\rm max} = 5 f_{\rm local}$, so the 
Nyquist frequency is $f_{\rm Nyquist} = 10 f_{\rm local}$. If the signal is sampled at 
the corresponding Nyquist time interval $\Delta T=1/f_{\rm Nyquist}=1/20$, 20 data 
points would be needed. If these 20 points are sampled uniformly in time, then the 
signal in the two oscillation cycles can be reconstructed. The reconstruction becomes 
quite challenging due to two factors: the limited availability of only six data points 
and their random distribution across the unit time interval. Consider points \#5 and 
\#6, which occur during a downward oscillation cycle in the ground truth data. 
Accurately reconstructing this downward oscillation presents a key challenge. When 
training data from the same target system is available, standard machine learning 
techniques can faithfully reconstruct the dynamics~\cite{yeo2019data}, as demonstrated 
in Fig.~\ref{fig:main_idea}(b). However, without access to training data from the 
target system, previous methods were unable to reconstruct the dynamics from such 
sparse observations. A related question is, after the reconstruction,
can the long-term dynamics or attractor of the system be predicted? We shall 
demonstrate that both the reconstruction and long-term prediction problems can 
be solved with hybrid machine learning, as schematically illustrated in 
Figs.~\ref{fig:main_idea}(d-f).

\begin{figure} [ht!]
\centering
\includegraphics[width=\linewidth]{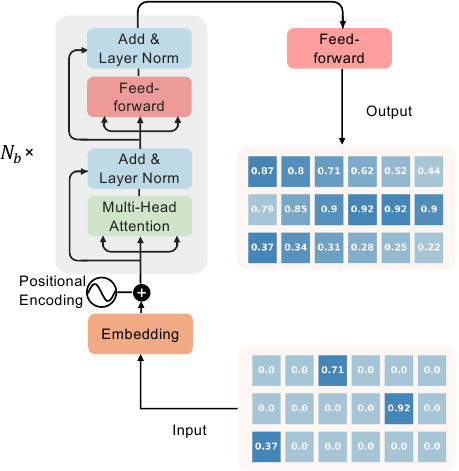} 
\caption{Transformer architecture. The transformer receives the sparse and random 
observation as the input and generates the reconstructed output. See text for a 
detailed mathematical description.} 
\label{fig:transformer}
\end{figure}

Consider a nonlinear dynamical system described by
\begin{align} \label{eq:dynamics}
\frac{d\mathbf{x}(t)}{dt} = \mathbf{f}(\mathbf{x}(t), t), \quad t \in [0, T],  
\end{align}
where $\mathbf{x}(t) \in \mathbb{R}^D$ is a $D$-dimensional state vector and 
$\mathbf{f}(\cdot)$ is the unknown velocity field. Let 
$\mathbf{X}=(\mathbf{x}_0,\cdots,\mathbf{x}_{L_s})^{\intercal}\in\mathbb{R}^{L_s\times D}$ 
be the full uniformly sampled data matrix of dimension $L_s\times D$ with each 
dimension of the original dynamical variable containing $L_s$ points. A sparse 
observational vector can be expressed as
\begin{align} \label{eq:obs}
\mathbf{\tilde{X}} = \mathbf{g}_{\alpha}(\mathbf{X}) (1 + \sigma \cdot \Xi),
\end{align}
where $\mathbf{\tilde{X}} \in \mathbb{R}^{L_s\times D}$ is the observational data
matrix of dimension $L_s\times D$ and $\mathbf{g}_{\alpha}(\cdot)$ is the following 
element-wise observation function:
\begin{align} \label{eq:obs_judge}
X^O_{ij} = g_{\alpha}(X_{ij}) = 
\begin{cases}
X_{ij},  & \quad \text{if} \ X_{ij} \ \text{is observed,} \\
0,  & \quad \text{otherwise,}
\end{cases}
\end{align}
with $\alpha$ representing the probability of matrix element $X_{ij}$ being observed. 
In Eq.~\eqref{eq:obs}, Gaussian white noise of amplitude $\sigma$ is present during 
the measurement process, where $\Xi \sim \mathcal{N}(0, 1)$. Our goal is utilizing 
machine learning to approximate the system dynamics function $\mathbf{f}(\cdot)$ by 
another function $\mathbf{f}'(\cdot)$, assuming that $\mathbf{f}$ is Lipschitz 
continuous with respect to $\mathbf{x}$ and the observation function produces sparse 
data: $\mathbf{g}: \mathbf{X} \rightarrow \mathbf{\tilde{X}}$. To achieve this, it is 
necessary to design a function $\mathcal{F}(\mathbf{\tilde{X}})=\mathbf{X}$ that 
comprises implicitly $\mathbf{f}'(\cdot)\approx \mathbf{f}(\cdot)$ so that it recovers 
the system dynamics by filling the gaps in the observation, where 
$\mathcal{F}(\mathbf{\tilde{X}})$ should have the capability of adapting to any 
given unknown dynamics. 

Selecting an appropriate neural network architecture for reconstructing dynamics 
from sparse data requires meeting two fundamental requirements: (1) dynamical memory 
to capture long-range dependencies in the sparse data, and (2) flexibility to handle 
input sequences of varying lengths. Transformers~\cite{vaswani2017attention}, 
originally developed for natural language processing, satisfy these requirements and 
have proven effective for time series analysis. Figure\ref{fig:transformer} 
illustrates the transformer's main structure. The data matrix $\mathbf{\tilde{X}}$ 
is first processed through a linear fully-connected layer with bias, transforming it 
into an $L_s\times N$ matrix. This output is then combined with a positional encoding 
matrix, which embeds temporal ordering information into the time series data.
This projection process can be described as~\cite{yildiz2022multivariate}:
\begin{align}
\mathbf{X}_p = \mathbf{\tilde{X}} \mathbf{W}_p  + \mathbf{W}_b + \mathbf{PE},
\end{align}
where $\mathbf{W}_p \in \mathbb{R}^{D\times N}$ represents the fully-connected layer 
with the bias matrix $\mathbf{W}_b \in \mathbb{R}^{L_s\times N}$ and the position 
encoding matrix is $\mathbf{PE} \in \mathbb{R}^{L_s\times N}$. Since the transformer 
model does not inherently capture the order of the input sequence, positional encoding 
is necessary to provide the information about the position of each time step. For a 
given position $1 \le {\rm pos} \le L_s^{max}$ and dimension $1 \le d \le D$, the 
encoding is given by
\begin{align}
	\mathbf{PE}_{pos,2d} = \sin{\left(\frac{{\rm pos}}{10000^{2d/N}}\right)},\\
	\mathbf{PE}_{pos,2d+1} = \cos{\left(\frac{{\rm pos}}{10000^{2d/N}}\right)},
\end{align}
The projected matrix $\mathbf{X}_p \in \mathbb{R}^{L_s\times N}$ then serves as the 
input sequence for $N_b$ attention blocks. Each block contains a multi-head attention 
layer, a residual layer (add \& layer norm), and a feed-forward layer, and a second 
residual layer. The core of the transformer lies in the self-attention mechanism, 
allowing the model to weight the significance of distinct time steps. The multi-head 
self-attention layer is composed of several independent attention blocks. The first 
block has three learnable weight matrices that linearly map $\mathbf{X}_p$ into query 
$\mathbf{Q}_1$ and key $\mathbf{K}_1$ of the dimension $L_s\times d_k$ and value 
$\mathbf{V}_1$ of the dimension $L_s\times d_v$: 
\begin{align}
\mathbf{Q}_1=\mathbf{X}_p\mathbf{W}_\mathbf{Q_1},\quad \mathbf{K}_1= \mathbf{X}_p\mathbf{W}_\mathbf{K_1}, \quad \mathbf{V}_1 = \mathbf{X}_p\mathbf{W}_\mathbf{V_1},
\end{align}
where $\mathbf{W}_{\mathbf{Q}_1} \in \mathbb{R}^{N\times d_k}$, 
$\mathbf{W}_{\mathbf{K}_1} \in \mathbb{R}^{N\times d_k}$, and 
$\mathbf{W}_{\mathbf{V}_1} \in \mathbb{R}^{N\times d_v}$ are the trainable weight 
matrices, $d_k$ is the dimension of the queries and keys, and $d_v$ is the dimension 
of the values. A convenient choice is $d_k=d_v=N$. The attention scores between the query 
$\mathbf{Q}_1$ and the key $\mathbf{K}_1$ are calculated by a scaled multiplication, 
followed by a softmax function:
\begin{align}
\mathbf{A}_{\mathbf{Q}_1, \mathbf{K}_1} = {\rm softmax} \left(\frac{\mathbf{Q}_1\mathbf{K}_1^\mathsf{T}}{\sqrt{d_k}}\right),
\end{align}
where $\mathbf{A}_{\mathbf{Q}_1,\mathbf{K}_1} \in \mathbb{R}^{L_s\times L_s}$. The softmax 
function normalizes the data with ${\rm softmax}(x_i) = \exp(x_i)/\sum_j \exp(x_j)$, 
and the $\sqrt{d_k}$ factor mitigates the enlargement of standard deviation due to 
matrix multiplication. For the first head (in the first block), the attention matrix is computed as a dot product 
between $\mathbf{A}_{\mathbf{Q}_1, \mathbf{K}_1}$ and $\mathbf{V}_1$:
\begin{align}
\mathbf{O}_{11} &= {\rm Attention}(\mathbf{Q}_1, \mathbf{K}_1, \mathbf{V}_1),\\ \nonumber
&= \mathbf{A}_{\mathbf{Q}_1, \mathbf{K}_1} \mathbf{V}_1 = {\rm softmax}\left(\frac{\mathbf{Q}_1\mathbf{K}_1^\mathsf{T}}{\sqrt{d_k}}\right) \mathbf{V}_1, 
\end{align}
where $\mathbf{O}_{11} \in \mathbb{R}^{L_s\times d_v}$. The transformer employs 
multiple ($h$) attention heads to capture information from different subspaces. The 
resulting attention heads $\mathbf{O}_{1i}$ ($i=1,\ldots,h$) are concatenated and 
mapped into a sequence $\mathbf{O}_1 \in \mathbb{R}^{L_s\times N}$, 
described as:
\begin{align}
\mathbf{O}_1 = \mathcal{C}(\mathbf{O}_{11}, \mathbf{O}_{12}, \cdots \mathbf{O}_{1h}) \mathbf{W}_{o1},
\end{align}
where $\mathcal{C}$ is the concatenation operation, $h$ is the number of heads, and 
$\mathbf{W}_{o1} \in \mathbb{R}^{h d_v\times N}$ is an additional matrix for linear 
transformation for performance enhancement. The output of the attention layer 
undergoes a residual connection and layer normalization, producing $\mathbf{X}_{R1}$ 
as follows:
\begin{align}
\mathbf{X}_{R1} = {\rm LayerNorm}(\mathbf{X_p}+{\rm Dropout}(\mathbf{O}_1))
\end{align}
A feed-forward layer then processes this data matrix, generating output 
$\mathbf{X}_{F1} \in \mathbb{R}^{L_s\times N}$ as:
\begin{align}
	\mathbf{X}_{F1} = \max (0, \mathbf{X}_{R1} \mathbf{W}_{F_a}+\mathbf{b}_a) \mathbf{W}_{F_b} + \mathbf{b}_b,
\end{align}
where $\mathbf{W}_{F_a} \in \mathbb{R}^{N \times d_f}$, 
$\mathbf{W}_{F_b} \in \mathbb{R}^{d_f \times N}$, $\mathbf{b}_a$ and $\mathbf{b}_b$ are 
biases, and $\max(0, \cdot)$ denotes a ReLU activation function. This output is again 
subjected to a residual connection and layer normalization. 

The output of the first block operation is used as the input to the second block. The 
same procedure is repeated for each of the remaining $N_b - 1$ blocks. The final output 
passes through a feed-forward layer to generate the prediction. Overall, the whole 
process can be represented as $\mathbf{Y} = \mathcal{F}(\mathbf{\tilde{X}})$. 

To evaluate the reliability of the generated output, we minimize a combined loss 
function with two components: (1) a mean squared error (MSE) loss that measures 
absolute error between the output and ground truth, and (2) a smoothness loss that 
ensures the output maintains appropriate continuity. The loss 
function is given by
\begin{align}
\mathcal{L} = \alpha_1 \mathcal{L}_{\rm mse} + \alpha_2 \mathcal{L}_{\rm smooth},
\end{align}
where $\alpha_1$ and $\alpha_2$ are scalar weights controlling the trade-off between 
the two loss terms. The first component $\mathcal{L}_{\rm mse}$ measures the absolute 
error between the predictions and the ground truth:
\begin{align} \label{eq:mse}
\mathcal{L}_{\rm mse} = \frac{1}{n}\sum_{i=1}^{n}(y_i-\hat{y}_i)^2,
\end{align}
with $n$ being the total number of data points, $y_i$ and $\hat{y}_i$ denoting the 
ground truth and predicted value at time point $i$, respectively. The second component 
$\mathcal{L}_{\rm smooth}$ of the loss function consists of two terms: Laplacian 
regularization and total variation regularization, which penalize the second-order 
differences and absolute differences, respectively, between consecutive predictions. 
The two terms are given by:
\begin{align}
\mathcal{L}_{\rm laplacian} = \frac{1}{n-2}\sum_{i=2}^{n-1}(\hat{y}_{i-1}+\hat{y}_{i+1}-2\hat{y})^2,
\end{align}
and
\begin{align}
\mathcal{L}_{\rm tv} = \frac{1}{n-1} \sum_{i=1}^{n-1} |\hat{y}_i-\hat{y}_{i+1}|.
\end{align}
We assign the same weights to the two penalties, so the final combined loss function 
to be minimized is
\begin{align}
\mathcal{L} = \mathcal{L}_{\rm mse} + \alpha_s (\mathcal{L}_{\rm laplacian} + \mathcal{L}_{\rm tv}).
\end{align}
We set $\alpha_s = 0.1$.

The second component of our hybrid machine-learning framework is reservoir computing,
which takes the output of the transformer as the input to reconstruct the long-term 
``climate'' or attractor of the target system. A detailed description of reservoir 
computing used in this context and its hyperparameters optimization are presented
in Appendices~\ref{appendix_b} and \ref{appendix_c}. 

\section{Results} \label{sec:results}

\begin{figure*} [ht!]
\centering
\includegraphics[width=\linewidth]{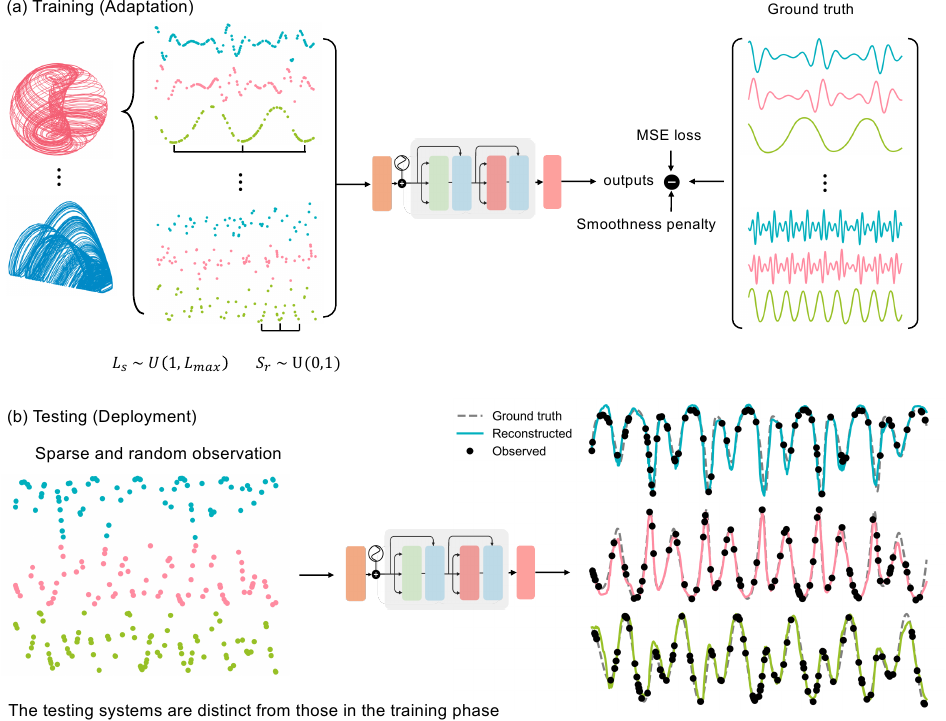} 
\caption{Illustration of the transformer-based dynamics reconstruction framework. 
(a) Training (adaptation) phase, where the model is trained on various synthetic chaotic 
systems, each divided into segments with randomly distributed sequence lengths $L_s$ and 
sparsity $S_r$. The data is masked before being input into the transformer, and the 
ground truth is used to minimize the MSE loss and smoothness loss with the output. By 
learning a randomly chosen segment from a random training system each time, the 
transformer is trained to handle data with varying lengths and different levels of 
sparsity. (b) Testing (deployment) phase. The testing systems are distinct from those 
in the training phase, i.e., the transformer is not trained on any of the testing 
systems. Given sparsely observed set of points, the transformer is able to reconstruct 
the dynamical trajectory.}
\label{fig:train_test}
\end{figure*}

\begin{table}[b]
\caption{\label{tab:hyper_t}
Optimal transformer hyperparameter values}
\begin{ruledtabular}
\begin{tabular}{cc}
Hyperparameters  & Descriptions \\
\hline 
\\
$D_l=1,500,000$ & Training length for each system \\
$d_f=512$ & Number of Feedforward neurons \\
$h=4$ & Number of Transformer heads \\
$N_b=4$ & Number of Transformer blocks \\
$N=128$ & input embedded dimension \\
$\sigma=0.05$ & Measurement noise level \\
$L_s^{\max}=3000$ & Maximum input sequence length \\
$b_s=16$ & Batch size \\
$\rm epoch =50$ & Epoch \\
$lr=0.001$ & Learning rate\\
$P_{drop}=0.2$ & Dropout rate \\ 
\end{tabular}
\end{ruledtabular}
\end{table}

We test our approach on three nonlinear dynamical systems in the deployment phase:
a three-species chaotic food chain system~\cite{mccann1994nonlinear}, the 
classic chaotic Lorenz system~\cite{lorenz1963deterministic}, and Lotka-Volterra 
system~\cite{vano2006chaos}. The transformer had no prior exposure to these systems 
during its training (adaptation) phase. We use sparse observational data from each 
system to reconstruct their underlying dynamics.

In the training phase, 28 synthetic chaotic systems are used to train the transformer, 
enabling it to learn to extract dynamic behaviors from sparse observations 
(see Appendix~\ref{appendix_a} for a detailed description). To enable the transformer
to handle data from new, unseen systems of arbitrary time series length $L_s$ and 
sparsity $S_r$, we employ the following strategy at each training step: (1) randomly 
selecting a system from the pool of synthetic chaotic systems and (2) preprocessing the 
data from the system using a uniformly distributed time series length 
$L_s\sim U(1,L_s^{\max})$, and uniformly distributed sparsity $S_r\sim U(0,1)$. By so 
doing, we prevent the transformer from learning any specific system dynamics too well, 
thereby encouraging it to treat each set of inputs as a new system. In addition, the 
strategy teaches the transformer to master as many features as possible. 
Figure~\ref{fig:train_test}(a) illustrates the training phase, with examples shown 
on the left side. On the right side, the sampled examples are encoded and fed into the 
transformer. The performance is evaluated by MSE loss and smoothness loss between the 
output and ground truth, and is used to update the neural-network weights. 

Unless otherwise stated, the following computational settings for machine learning 
are used. Given a target system, time series are generated numerically by integrating 
the system with time step $dt=0.01$. The initial states of both the dynamical process 
and the neural network are randomly set from a uniform distribution. An initial phase 
of the time series is removed to ensure that the trajectory has reached the attractor. 
The training and testing data are obtained by sampling the time series at the interval 
$\Delta_s$ chosen to ensure an acceptable generation. Specifically, for the chaotic 
food chain, Lorenz and Lotka-Volterra systems, we set $\Delta_s=10dt=0.1$, corresponding 
to approximately 1 over $40\sim 50$ cycles of oscillation. A similar procedure is also 
applied to other synthetic chaotic systems. The time series data are preprocessed by 
using min-max normalization so that they are in the range [0,1]. The complete data 
length for each system is 1,500,000 (about 30,000 cycles of oscillations), which is 
divided into segments with randomly chosen sequence lengths $L_s$ and sparsity $S_r$. 
For the transformer, we use a maximum sequence length of 3,000 (corresponding to about
60 cycles of oscillations) - the limitation of input time series length. We apply 
Bayesian optimization~\cite{bayesian} and a random search 
algorithm~\cite{bergstra2012random} to systematically explore and identify the optimal 
set of various hyperparameters. Two chaotic Sprott systems - $\rm Sprott_0$ and 
$\rm Sprott_1$ - are used to find the optimal hyperparameters, ensuring no data 
leakage from the testing systems. The optimized hyperparameters for the transformer are 
listed in Table~\ref{tab:hyper_t}. Simulations are run using Python on computers with 
six RTX A6000 NVIDIA GPUs.

For clarity of presentation, in the main text, we focus on the results from the chaotic 
three-species food-chain system, while leaving the results from the other two testing
systems in Appendix~\ref{appendix_d}.

\begin{figure*} [ht!]
\centering
\includegraphics[width=\linewidth]{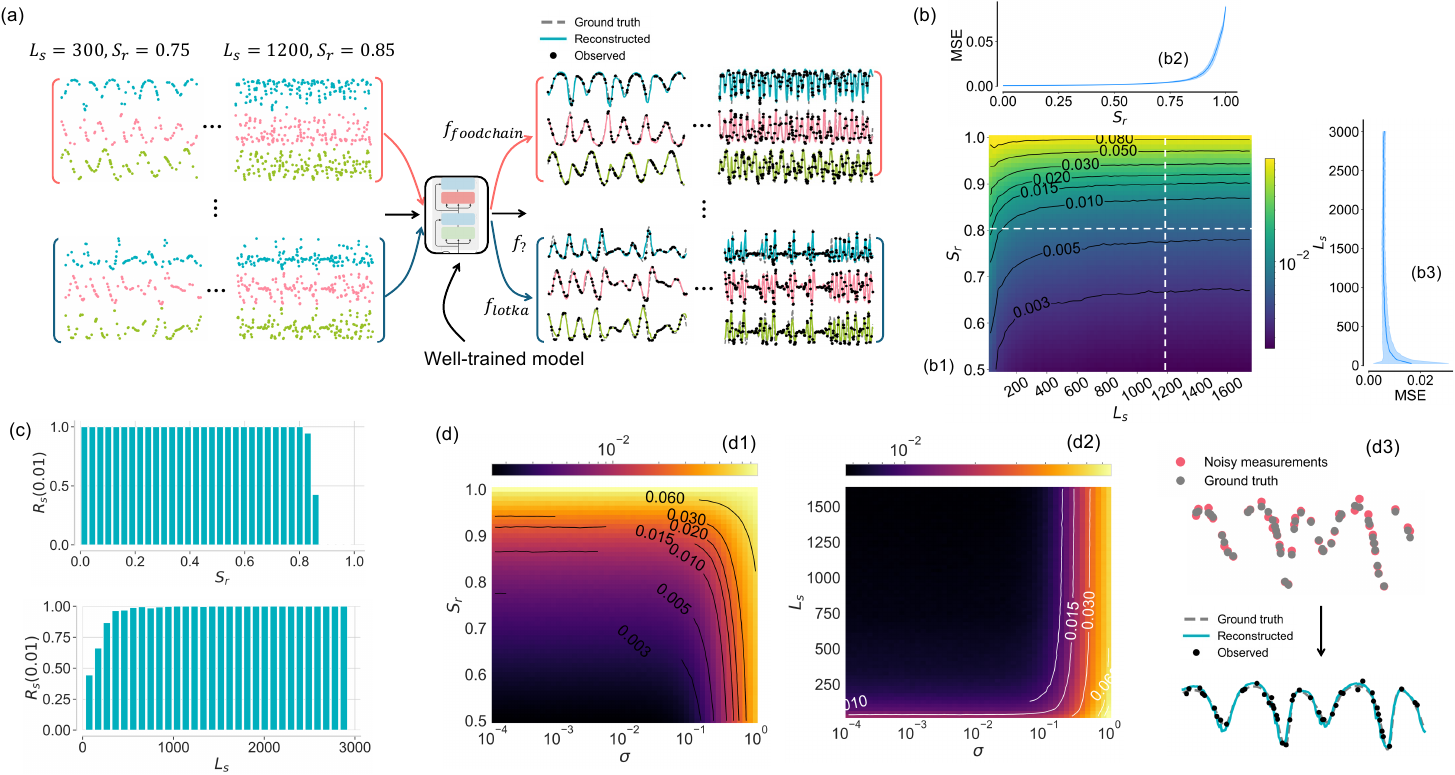} 
\caption{Performance of dynamics reconstruction. (a) Illustration of reconstruction 
results for the chaotic food-chain and Lotka-Volterra systems as the testing targets 
that the transformer has never been exposed to. For each target system, two sets of 
sparse measurements of different length $L_s$ and sparsity $S_r$ are shown. The 
trained transformer reconstructs the complete time series in each case. (b) Color-coded 
ensemble-averaged MSE values in the parameter plane $(L_s,S_r)$ (b1). Examples of 
testing MSE versus $S_r$ and $L_s$ only are shown in (b2) and (b3), respectively. 
(c) Ensemble-averaged reconstruction stability indicator $R_s({\rm MSE_c})$ versus 
$S_r$ and $L_s$, the threshold MSE is $\rm MSE_c=0.01$. (d) Robustness of dynamics 
reconstruction against noise: ensemble-averaged MSE in the parameter plane 
$(\sigma,S_r)$ (d1) and $(\sigma,L_s)$ (d2), with $\sigma$ being the noise amplitude. 
An example of reconstruction under noise of amplitude $\sigma=0.1$ is shown in (d3). 
The values of the performance indicators are the result of averaging over 50 
independent statistical realizations.}
\label{fig:statistics}
\end{figure*}

\subsection{Dynamics reconstruction}

The three species food-chain system~\cite{mccann1994nonlinear} is described by 
\begin{align}\label{eq:foodchain}
\frac{dR}{dt} &= {R(1-\frac{R}{\rm K}) - \frac{ {\rm x_c y_c} C R}{R+ {\rm R_0}}}, \nonumber \\
\frac{d C}{dt} &= {\rm x_c} C (\frac{{\rm y_c} R}{R+{\rm R_0}}-1) - \frac{{\rm x_p y_p } P C}{C+{\rm C_0}} , \\
\frac{d P}{dt} &= {\rm x_p} P(\frac{ {\rm y_p} C}{C + {\rm C_0}}-1), \nonumber
\end{align}
where $R$, $C$, and $P$ are the population densities of the resource, consumer,
and predator species, respectively. The system has seven parameters:
${\rm K,x_c,y_c,x_p,y_p,R_0,C_0}>0$. Figure~\ref{fig:train_test}(b) presents an
example of recovering the dynamics of the chaotic food-chain system for $L_s=2000$ 
and $S_r=0.85$. The target output time series for each dimension should contain
2000 points (around 40 cycles of oscillations), but only randomly selected 
$L_s \cdot (1-S_r) = 300$ points are exposed to the trained transformer. The right side 
of Fig.~\ref{fig:train_test}(b) shows the recovered time series, where the three 
dynamical variables are represented by different colors, the black points indicate 
observations, and the grey dashed lines are the ground truth. Only a segment of a 
quarter of the points is displayed. This example demonstrates that, with such a high 
level of sparsity, directly connecting the observational points will lead to significant 
errors. Instead, the transformer infers the dynamics by filling the gaps with the 
correct dynamical behavior. It is worth emphasizing that the testing system has never 
been exposed to the transformer during the training phase, requiring the neural machine 
to explore the underlying unknown dynamics from sparse observations based on experience 
learned from other systems. Extensive results with varying values of the parameters
$L_s$ and $S_r$ for the three testing systems can be found in Appendix~\ref{appendix_d}.

\begin{figure*} [ht!]
\centering
\includegraphics[width=\linewidth]{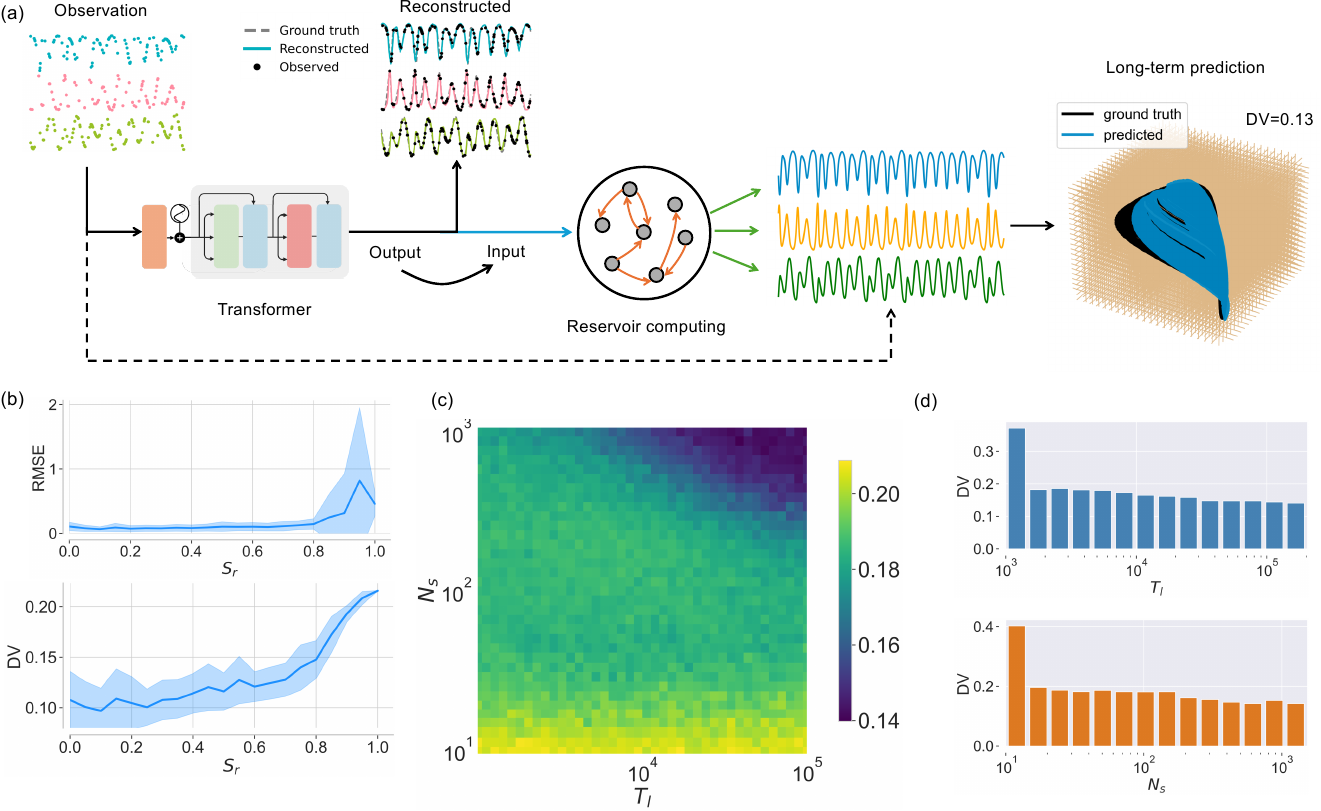} 
\caption{Reservoir-computing based long-term dynamics prediction. (a) An illustration 
of hybrid transformer/reservoir-computing framework. The time series reconstructed 
by the transformer is used to train the reservoir computer that generates time series 
of the target system of arbitrary length, leading to a reconstructed attractor
that agrees with the ground truth. (b) RMSE and DV versus the sparsity parameter. 
(c) Color-coded ensemble-averaged DV in the reservoir-computing hyperparameter plane 
$(T_l,N_s)$ for $S_r=0.8$. (d) DV versus training length $T_l$ for $N_s = 500$ and
versus reservoir network size $N_s$ for $T_l=10^5$. In all cases, 50 independent 
realizations are used.} 
\label{fig:rc}
\end{figure*}

\subsection{Performance of dynamics reconstruction}

To characterize the performance of dynamics reconstruction, we use two measures: MSE 
[Eq.~\eqref{eq:mse}] and prediction stability $R_s({\rm MSE_c})$, the probability 
that the transformer generates stable predictions, defined as the probability that
the MSE is below a predefined stable threshold $\rm MSE_c$:
\begin{align} \label{eq:recovery_stability}
R_s({\rm MSE_c}) = \frac{1}{n}\sum_{i=1}^n [{\rm MSE < MSE_c}],
\end{align}
where $n$ is the number of iterations and $[\cdot]=1$ if the statement inside is true 
and zero otherwise. Figure~\ref{fig:statistics}(a) shows the working of the framework 
in the testing phase: a well-trained transformer receives inputs from previously 
unseen systems, with random sequence length $L_s$ and sparsity $S_r$, and is able to
reconstruct the dynamics. Some representative time-series segments of the reconstruction 
and the ground truth are displayed. Figures~\ref{fig:statistics}(b) and 
\ref{fig:statistics}(c) depict the ensemble-averaged reconstruction performance for the 
chaotic food chain system. As $L_s$ increases and $S_r$ decreases, the transformer can 
gain more information to facilitate reconstructions. When the available data become 
more sparse, the performance degrades. Overall, for $S_r < 0.8$ and a recovered sequence 
length larger than 500 (about 10 cycles of oscillations in the data), satisfactory 
reconstruction can be achieved.

It is useful to study how noise affects the dynamics reconstruction. We use normally 
distributed stochastic processes of zero mean and standard deviation $\sigma$,
which perturbs the observational points $x$ to $x + x\cdot \xi$ after normalization.
This is multiplicative or demographic noise commonly encountered in ecological systems.
Figure~\ref{fig:statistics}(d) shows the effects of the noise on the reconstruction
performance, where 50 independent runs are used to calculate the average MSE. The 
results indicate that, for reasonably small noise (e.g., $\sigma < 10^{-1}$), robust 
reconstruction with relatively low MSE values can be achieved. We have also studied
the effect of additive noise, with results presented in Appendix~\ref{appendix_e}. 

\subsection{Prediction of long-term dynamical climate}

The results presented so far are for reconstruction of relatively short term 
dynamics, where the sequence length $L_s$ is limited to below 3000, corresponding 
to approximately 60 cycles of dynamical oscillations in the data. Can the long-term
dynamical behavior or climate as characterized by, e.g., a chaotic attractor, be 
faithfully predicted? To address this question, we note that reservoir computing has 
the demonstrated ability to generate the long-term behavior of chaotic 
systems~\cite{FJZWL:2020,KFGL:2021a,KWGHL:2023,zhai2023emergence,kong2024reservoir}.
Our solution is then employing reservoir computing to learn the output time series 
generated by the transformer so as to further process the reconstructed time series. 
Assume that a number of sparse data segments are available. The corresponding 
transformer-reconstructed segments are then used as the training data for the 
reservoir computer for it to find the relationship between the dynamical state at the
current step and that in the immediate future. The trained reservoir computer can 
predict or generate any length of time series of the target system, as exemplified 
in Fig.~\ref{fig:rc}(a). It can be seen that the reservoir-computing generated 
attractor agrees with the ground truth. More details about reservoir computing, its 
training and testing can be found in Appendix~\ref{appendix_d}. 

To evaluate the performance of the reservoir-computing generated long-term dynamics, 
we use two measures: root MSE (RMSE) and deviation value (DV) characterizing the 
topological distance between the predicted attractor and the ground 
truth~\cite{zhai2023emergence}. Specifically, for a three-dimensional target system,
we divide the three-dimensional phase space into a uniform cubic lattice with the 
cell size $\Delta=0.05$ and count the number of trajectory points in each cell, for 
both the predicted and true attractors in a fixed time interval. The DV measure is
defined as
\begin{align}\label{eq:DV}
{\rm DV} \equiv \sum_{i=1}^{m_x}\sum_{j=1}^{m_y}\sum_{k=1}^{m_z} \sqrt{(f_{i,j,k} - \hat{f}_{i,j,k})^2},
\end{align}
where $m_x$, $m_y$, and $m_z$ are the total numbers of cells in the $x$, $y$, and $z$ 
directions, respectively, $f_{i,j,k}$ and $\hat{f}_{i,j,k}$ are the frequencies of 
visit to the cell $(i,j,k)$ by the predicted and true trajectories, respectively. If 
the predicted trajectory leaves the phase space boundary, we count it as if it has landed 
in the boundary cells where the true trajectory never goes. Figure~\ref{fig:rc} (b) 
presents the short- and long-term prediction performance by comparing the 
reservoir-computing predicted attractors with the ground truth. We calculate the RMSE 
using a short-term prediction length of 150 (approximately 3 cycles of oscillations), 
and the DV using a long-term prediction length of 10,000 (corresponding approximately 
to 200 cycles of oscillations). It can be seen that the reconstructed time series and 
attractors are close to their respective ground truth when the sparsity parameter $S_r$ 
is below 0.8, as indicated by the low RMSE and DV values. The number of available data 
segments from the target system tends to have a significant effect on the prediction 
accuracy. Figures~\ref{fig:rc}(c) and \ref{fig:rc}(d) show the dependence of the DV on 
two reservoir-computing hyperparameters: the training length $T_l$ and the reservoir 
network size $N_s$. As the training length and network size increase, DV decreases, 
indicating improved performance.

\section{Discussion} \label{sec:discussion}

Exploiting machine learning to understand, predict and control the behaviors of 
nonlinear dynamical systems have demonstrated remarkable success in solving previously
deemed difficult problems~\cite{zhai2023model,kim2023neural,zhai2024parameter}. 
However, an essential prerequisite for these machine-learning studies is the 
availability of training data. Often, extensive and uniformly sampled data of the 
target system are required for training. In addition, in most previous works, training
and testing data are from the same system, with a focus on minimizing the average 
training errors on the specific system and greedily improving the performance by 
incorporating all correlations within the data, the so-called iid (independently
and identically distributed) assumption. While the iid setting can be effective, 
unforeseen distribution shifts during testing or deployment can cause the optimization 
purely based on the average training errors to perform poorly~\cite{liu2021towards}. 
Several strategies have been proposed to handle nonlinear dynamical systems. One 
approach trains neural networks using data from the same system under different 
parameter regimes, enabling prediction of new dynamical behaviors including critical 
transitions~\cite{KFGL:2021a}. Another method uses data from multiple systems to 
train neural networks in tasks like memorizing and retrieving complex dynamical 
patterns~\cite{kong2024reservoir,du2024multi}. However, this latter approach fails 
when encountering novel systems not present in the training data. Meta-learning has been 
shown to achieve satisfactory performance with only limited data, but training data 
from the target systems are still required to fine-tune the network 
weights~\cite{zhai2024learning}. In addition, a quite recent work used well-defined, 
pre-trained large language models not trained using any chaotic data and showed that 
these models can predict the short-term and long-term dynamics of chaotic 
systems~\cite{zhang2024zero}. 

We developed a hybrid machine-learning framework to construct the dynamics of  
target systems, under two limitations: (1) the available observational data are 
sparse and (2) no training data from the system are available. This requires that 
the machine-learning framework already be well trained before deploying to the target 
system. We address this challenge by training the transformer using synthetic or 
simulated data from numerous chaotic systems, completely excluding data from the 
target system. This allows direct application to the target system without 
fine-tuning. To ensure the transformer's effectiveness on previously unseen systems, 
we implement a ``triple-randomness'' training regime that varies the training systems, 
input sequence length, and sparsity level. As a result, the
transformer will treat each dataset as a new system, rather than adequately learning 
the dynamics of any single training system. Because of the synthetic nature of the 
training systems, a massive dataset for each training system is used. This process 
continues with data from different chaotic systems with random input sequence length 
and sparsity until the transformer is well-experienced and able to ``perceive'' the 
underlying dynamics from the sparse observations. The end result of this training 
process is that the transformer gains ``knowledge'' through its experience by adapting 
to the diverse synthetic datasets. During the testing or deployment phase, the 
transformer reconstructs dynamics from sparse data of arbitrary length and sparsity 
drawn from a completely new dynamical system. When multiple segments of sparse 
observations are available, we can reconstruct the system's long-term `climate' 
through a two-step process. First, the transformer repeatedly reconstructs system 
dynamics from these data segments. Second, reservoir computing uses these 
transformer-reconstructed dynamics as training data to generate system evolution 
over any time duration. This hybrid approach, combining transformer and reservoir 
computing methods, enables reconstruction of the target system's long-term dynamics 
and attractor from sparse data alone. The combination of the transformer and 
reservoir computing constitutes our hybrid machine-learning framework. 

We emphasize the key feature of our hybrid framework: recovering the dynamics from 
sparse observations of an unseen dynamical system, even when the available data has a 
high degree of sparsity. We have tested the framework on two benchmark ecosystems and 
one classical chaotic system. In all cases, with extensive training conducted on 
synthetic datasets under diverse settings, accurate and robust reconstruction has
been achieved. Empirically, the minimum requirements for the transformer to be 
effective are: the dataset from the target system should have the length of at least 
20 average cycles of its natural oscillation and the sparsity degree is less than 80\%. 
For the subsequent reservoir computing learning, at least three segments of the time 
series data from the transformer are needed for reconstructing the attractor of the 
target system. We have also addressed issues including the effect of noise and 
hyperparameter optimization. The key to the success of the hybrid framework lies in 
{\em versatile dynamics}: with training based on the dynamical data from a diverse 
array of synthetic systems, the transformer will gain the ability to reconstruct the 
dynamics of ``never-seen'' target systems. We have provided a counter example that, 
when there are no dynamics in the time series, the framework fails to perform the 
reconstruction task (see Appendix~\ref{appendix_f}). 

Our hybrid transformer/reservoir-computing framework represents a powerful tool for 
dynamics reconstruction and prediction of long-term behavior in situations where only 
sparse observations from a newly encountered system are available. In fact, such a 
situation is expected to arise in different fields. For example, in ecosystems, the 
available data are often limited and incomplete - the reason that we employed two 
ecosystems as our testing examples. Possible applications extend to medical and 
biological systems, particularly in wearable health monitoring where data collection 
is often interrupted. For instance, smartwatches and fitness trackers regularly 
experience gaps due to charging, device removal during activities like swimming, 
or signal interference. Another potential application is predicting critical 
transitions from sparse and noisy observations, such as detecting when an athlete's 
performance metrics indicate approaching overtraining, or when a patient's vital 
signs suggest an impending health event. In these cases, our hybrid framework can 
reconstruct complete time series from incomplete wearable device data, serving as 
input to parameter-adaptable reservoir computing~\cite{KFGL:2021a,KWGHL:2023,PL:2024} 
for anticipating these critical transitions. This approach is particularly valuable 
for continuous health monitoring where data gaps are inevitable, whether from smart 
devices being charged, removed, or experiencing connectivity issues.

\section*{Acknowledgment}

This work was supported by the Air Force Office of Scientific Research under Grant 
No.~FA9550-21-1-0438 and by the Office of Naval Research under Grant No.~N00014-24-1-2548.

\appendix

\section{Additional background} \label{appendix_na}

For dynamical systems reconstruction, machine-learning methods are basically complex 
approximators mapping inputs to the outputs, or fit relationships among the variables. 
Deep learning, in particular, is a purely data-driven method that relies on extensive 
datasets to discover the input-output relationships. This process inherently causes the 
model to overfit to a specific domain distributed about the training data. As a result, 
models trained on one dataset often struggle to adapt to unseen data from new 
distributions, a phenomenon known as out-of-distribution generalization or distribution 
(covariate) shift~\cite{yu2024learning}. This limitation is particularly pronounced 
when learning nonlinear dynamics. The challenge is compounded not only by the 
nonstationary nature of the dynamics but also by changes in the system parameters, 
which can cause distribution shifts. More extreme shifts can occur when switching 
between different systems.

Recent years have witnessed a growth of interest in machine learning methods leading to 
predictive frameworks to estimate data values using unsupervised or supervised learning,
due to their nonlinearity, flexibility, and ability to capture useful information 
embedded in the observed data. Traditional methods include K-nearest neighbor 
(KNN)~\cite{batista2002study}, support vector machine (SVM)~\cite{pelckmans2005handling}, 
and MissForest~\cite{stekhoven2012missforest}, etc., but deep learning models are able to
deliver more accurate predictions, which include three major categories: recurrent neural 
network (RNN)-based, generative models, and self-attention based~\cite{du2023saits}. 
RNN-based methods~\cite{che2018recurrent}, featured by their recurrent neural-network 
structure, are time-consuming and memory-constrained, making it difficult to capture 
the long-term dependencies within the time series. In addition, the RNN methods are 
commonly iterative, which inevitably introduces compounding errors through the process. 
Generative models include generative adversarial network (GAN)~\cite{yoon2018gain}, 
variational autoencoder (VAE)~\cite{ramchandran2021longitudinal}, and diffusion 
models~\cite{wang2023observed}. For example, the GAN-based method GAIN~\cite{yoon2018gain} 
takes a generator and a discriminator to predict the data values, where the former 
imputes the missing values conditioned on the observed data and outputs a reconstructed 
complete time series, and the latter determines which values are predicted and which 
belong to the observed data. Generative models provide an adversarial point of view to 
improve the performance of the generator, but it may suffer from non-convergence or 
mode collapse due to the loss formulation, and thus are difficult to 
train~\cite{salimans2016improved}. Self-attention-based methods, e.g., transformer 
models~\cite{vaswani2017attention}, focus on capturing the long-range dependencies and 
complex relationships within the time series. They excel at handling multi-dimensional 
data and at tackling input and output sequences of arbitrary length~\cite{liu2023one}. 
However, transformer models require substantial amounts of data for effective training
and for avoiding overfitting. For instance, a vision transformer (ViT) usually yields 
low accuracies on mid-size datasets but attains high accuracies when trained on 
sufficiently large datasets (e.g., 14M-300M images)~\cite{dosovitskiy2020image}. 
Transformer model is thus preferable when the resources are adequate due to its 
computational efficiency and scalability. Overall, most previous works focused on 
predicting unobserved values by assembling observable points in various ways, while 
marginalizing the underlying dynamics, which can be effective for specialized 
applications but is difficult to be generalized to complex dynamical systems.

It is worth mentioning the related problem of missing data imputation. 
Handling missing data can be approached through deletion or imputation. Deletion entails 
removing all entries with missing values~\cite{emmanuel2021survey}. This method is simple 
and effective when the missing data rate is low, typically less than 10$\%$ or 15$\%$, so 
the removal procedure does not significantly affect the 
analysis~\cite{strike2001software}. However, as the missing rate increases, the deletion 
method becomes less effective and can lead to biased 
conclusions~\cite{acuna2004treatment}. 
In contrast, imputation replaces missing data with estimated values using statistical or 
machine learning techniques~\cite{lin2020missing}. Simple imputation methods, such as 
replacing missing values with the mean or median of the available values, can be readily
implemented and are often used during data preprocessing. However, these methods can 
produce biased or unrealistic results, especially for high-dimensional datasets. 
Regression is another common imputation technique, where missing values are predicted 
using a regression model built from complete observations~\cite{song2007missing}. While 
effective, this method requires a large amount of data and does not account for any 
inherent variability. A more advanced approach is the Bayesian method, which treats 
missing values as unknown parameters drawn from an appropriate probability 
distribution~\cite{zhou2024review}. This approach allows for the incorporation of prior 
knowledge about the data distribution and specifies a probabilistic model that captures 
the relationship between the observed and missing values. 

\section{Adaptation (training) and deployment (testing) systems} \label{appendix_a}

\begin{figure} [ht!]
\centering
\includegraphics[width=\linewidth]{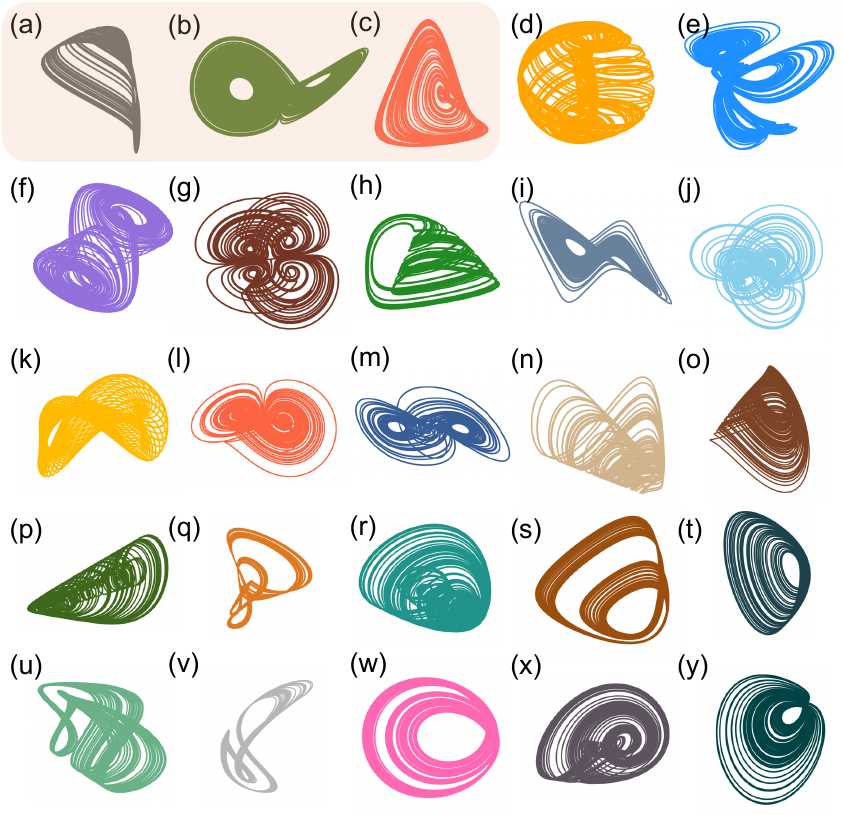} 
\caption{The chaotic systems for adaptation learning during the training and the 
target systems for testing. (a-c) Target testing systems: food-chain, Lorenz, and 
Lota-Volterra systems, respectively. Systems for learning: (d) Aizawa, (e) Bouali, 
(f) Chua, (g) Four wing, (h) Hastings-Powell, (i) Rikitake, (j) Wang, and
(k-y) Sprott systems.}
\label{fig:chaotic_systems}
\end{figure}

We provide the details of the systems employed during both the adaptation and 
deployment phases of our hybrid machine-learning framework as in 
Fig.~\ref{fig:chaotic_systems}, some of which are collected from~\cite{zhai2024learning}. The equations and parameters used to simulate the
time series are listed in Tabs.~\ref{tab:chaos} and \ref{tab:sprott}. For the
target testing systems, in addition to the three-species food chain system 
discussed in detail in the main text, two other systems are tested: the classic 
chaotic Lorenz system and the Lotka-Volterra system.
The Lorenz system is described by
\begin{align}
\frac{dx}{dt}&=\sigma_{lo} (y-x), \nonumber \\
\frac{dy}{dt}&=x (\rho_{lo}-z) - y, \\
\frac{dz}{dt}&=xy - \beta_{lo} z, \nonumber
\end{align}
where $\sigma_{lo}=10$, $\rho_{lo}=2.67$, and $\beta_{lo}=26$ are parameters.
The Lotka-Volterra system was originated from the modeling of predator-prey 
interactions and of certain chemical reactions~\cite{vano2006chaos}. For a system of 
$N_c$ species with population $P_i$ ($i=1,\ldots, N_c$) competing for finite resources,
the equations are given by
\begin{align}
\frac{dP_i}{d} = r_i P_i (1 - \sum_{j=1}^{N_c} a_{ij} P_j),
\end{align}
where $r_i$ is the growth rate of species $i$ and $a_{ij}$ represents characterizes
the interaction between species $j$ and $i$. We use the four-species Lotka-Volterra 
model~\cite{vano2006chaos}, with the parameter values:
\begin{align}
r_i = \begin{bmatrix}
       1  \\[0.3em]
       0.72 \\[0.3em]
       1.53 \\[0.3em]
       1.27
     \end{bmatrix},
\quad 
a_{ij} = \begin{bmatrix}
       1 & 1.09 & 1.52 & 0    \\[0.3em]
       0 & 1    & 0.44 & 1.36 \\[0.3em]
       2.33 & 0 & 1    & 0.47 \\[0.3em]
       1.21 & 0.51 & 0.35 & 1
     \end{bmatrix},
\end{align}
While transformer has the advantage of flexible input data length, the dimension of the
input vector needs to be fixed. Since the trained systems are three-dimensional, the 
target systems should have the same dimension. For the generated four-dimensional 
Lotka-Volterra time series, we take only data from the first three variables as the 
target. Since this operation discards the information of the fourth dynamical variable,
the reconstruction task becomes even more challenging. 

\begin{table*}[b]
\caption{\label{tab:chaos}
Chaotic systems}
\begin{ruledtabular}
\begin{tabular}{ccc}
Systems & Equations & Parameters\\
\hline
\\
Aizawa & $\dot{x}=(z-b)x-dy$ & $a=0.95, b=0.7, c=0.6$ \\
& $\dot{y}=dx+(z-b)y$ & $d=3.5, e=0.25, f=0.1$ \\
& $\dot{z}=c+az-z^3/3-(x^2+y^2)(1+e^z)+fzx^3$ \\
\\
Bouali & $\dot{x}=x(a-y)+\alpha z$ & $\alpha=0.3, \beta=0.05$ \\
& $\dot{y}=-y(b-x^2)$ & $a=4, b=1, c=1.5, s=1$ \\
& $\dot{z}=-x(c-sz)-\beta z$\\
\\
Chua & $\dot{x}=\alpha(y-x-ht)$ & $\alpha=15.6, \gamma=1, \beta=28$\\
& $\dot{y}=\gamma(x-y+z)$ & $\mu_0=-1.143, \mu_1=-0.714$\\
& $\dot{z}=-\beta y$ & $ht=\mu_1 x + 0.5(\mu_0-\mu_1)(|x+1|-|x-1|)$\\
\\
Dadras & $\dot{x}=y-ax+byz$ & $a=3, b=2.7$\\
& $\dot{y}=cy-xz+z$ & $c=1.7, d=2, e=9$\\
& $\dot{z}=dxy-ez$  \\
\\

Four\: wing & $\dot{x}=ax+yz$ & $a=0.2, b=0.01, c=-0.4$\\
& $\dot{y}=bx+cy-xz$ \\
& $\dot{z}=-z-xy$ \\
\\
Hastings-Powell & $\dot{V}=V(1-V)-a_1 VH/ (b_1 V + 1)$ & $a_1=5, a_2=0.1$ \\
& $\dot{H}=a_1VH/(b_1 V + 1) - a_2 H P/ (b_2 H + 1) - d_1 H$ & $b_1=3, b_2=2$ \\
& $\dot{P}=a_2 H P/ (b_2 H + 1) - d_2 P$ & $d_1=0.4, d_2=0.01$ \\
\\
Rikitake & $\dot{x}=-\mu x + zy$ & \\
& $\dot{y}=-\mu y + x(z-a)$ & $\mu=2, a=5$\\
& $\dot{z}=1-xy$ & \\
\\
Rossler & $\dot{x}=-(y+z)$ & \\
& $\dot{y}=x+ay$ & $a=0.2, b=0.2, c=5.7$\\
& $\dot{z}=b+z(x-c)$ & \\
\\
Wang & $\dot{x}= x-yz$ & \\
& $\dot{y}=x-y+xz$ & $a=3$\\
& $\dot{z}=-az + xy$ & \\
\\
\end{tabular}
\end{ruledtabular}
\end{table*}

\begin{table}[b]
\caption{\label{tab:sprott}
Chaotic sprott systems}
\begin{ruledtabular}
\begin{tabular}{ccc}
Case & Equations &\\
\hline 
\\
0 & $\dot{x}=y$, $\; \dot{y}=-x + yz$, $\; \dot{z}=1-y^2$ \\
\\
1 & $\dot{x}=yz$, $\;\dot{y}=x-y$, $\;\dot{z}=1-xy$ \\
\\
2 & $\dot{x}=yz$, $\;\dot{y}=x-y$, $\;\dot{z}=1-x^2$ \\
\\
3 & $\dot{x}=-y$, $\;\dot{y}=x+z$, $\;\dot{z}=xz+3y^2$ \\
\\
4 & $\dot{x}=yz$, $\;\dot{y}=x^2-y$, $\;\dot{z}=1-4x$ \\
\\
5 & $\dot{x}=y+z$, $\;\dot{y}=-x+0.5y$, $\;\dot{z}=x^2-z$ \\
\\
6 & $\dot{x}=0.4x + z$, $\;\dot{z}=xz-y$, $\;\dot{y}=-x+y$ \\
\\
7 & $\dot{x}=-y + z^2$, $\;\dot{y}=x + 0.5y$, $\;\dot{z}=xz$ \\
\\
8 & $\dot{x}=-0.2y$, $\;\dot{y}=x + z$, $\;\dot{z}=x + y^2 - z$ \\
\\
9 & $\dot{x}=2z$, $\;\dot{y}=-2y + z$, $\;\dot{z}=-x+y+y^2$ \\
\\
10 & $\dot{x}=xy - z$, $\;\dot{y}=x - y$, $\;\dot{z}=x + 0.3z$ \\
\\
11 & $\dot{x}=y + 3.9z$, $\;\dot{y}=0.9x^2 - y$, $\;\dot{z}=1 - x$ \\
\\
12 & $\dot{x}=-z$, $\;\dot{y}=-x^2 - y$, $\;\dot{z}=1.7 + 1.7x + y$ \\
\\
13 & $\dot{x}=-2y$, $\;\dot{y}=x + z^2$, $\;\dot{z}=1 + y - 2z$ \\
\\
14 & $\dot{x}=y$, $\;\dot{y}=x - z$, $\;\dot{z}=x + xz + 2.7y$ \\
\\
15 & $\dot{x}=2.7y + z$, $\;\dot{y}=-x + y^2$, $\;\dot{z}=x + y$ \\
\\
16 & $\dot{x}=-z$, $\;\dot{y}=x - y$, $\;\dot{z}=3.1x + y^2 + 0.5z$ \\
\\
17 & $\dot{x}=0.9 - y$, $\;\dot{y}=0.4 + z$, $\;\dot{z}=xy - z$ \\
\\
18 & $\dot{x}=-x - 4y$, $\;\dot{y}=x + z^2$, $\;\dot{z}=1 + x$ \\
\end{tabular}
\end{ruledtabular}
\end{table}

\section{Reservoir computing} \label{appendix_b}

\begin{figure} [ht!]
\centering
\includegraphics[width=\linewidth]{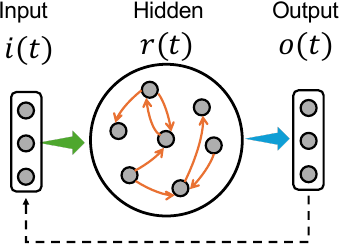} 
\caption{Basic configuration of reservoir computing.}
\label{fig:reservoir_computing}
\end{figure}

Reservoir computing, a class of recurrent neural networks, is a computational 
framework designed for efficient training and dynamical systems modeling. It 
consists of three layers: an input layer, a hidden recurrent layer, and an output 
layer. Unlike traditional recurrent neural networks where the weights associated
all layers are trained, in reservoir computing the input weights and internal 
recurrent connections remain fixed after random initialization, while only the 
output weights are adjusted during training through a linear regression. This 
setting enables efficient training with performance similar to that of the 
conventional recurrent neural networks~\cite{vlachas2020backpropagation}.

Figure~\ref{fig:reservoir_computing} illustrates the basic structure of reservoir
computing, where the input signal $\textbf{i}(t)$, typically a low-dimensional 
vector, is mapped into the high-dimensional state space in the hidden layer with 
$N_s$ nodes by the input matrix $\mathcal{W}_{\rm in}$. Activated by the sequence 
of reservoir input signals $[\mathbf{i}(:,1),\mathbf{i}(:,2),\cdots,\mathbf{i}(:,t)]$,
the hidden layer state $\mathbf{r}(t)$ is updated step-by-step according to
\begin{align} \label{eq:reservoir}
\mathbf{r}(t+1) &= (1-\alpha_r) \cdot \mathbf{r}(t) + \nonumber \\
&~~~~\alpha_r \cdot \tanh \left[\mathcal{A} \cdot \mathbf{r}(t)+\mathcal{W}_{\rm in}\cdot \mathbf{i}(t)\right],
\end{align}
where $\alpha_r$ is the leaking parameter that controls the speed of reservoir 
memory decays and the nonlinear activation function is of the hyperbolic tangent 
type. The dimensions of the input signal $\textbf{i}(t)$, hidden state \textbf{r}(t), 
and output signal \textbf{o}(t) are denoted as $D_i$, $N_s$, and $D_o$, respectively. 
For three-dimensional chaotic systems, we have $D_o=D_i=3$. The elements of the 
input matrix $\mathcal{W}_{\rm in}$, which has the dimension of $N_s*D_i$, are 
generated uniformly in the range $[-\gamma_r,\gamma_r]$ prior to training. The 
elements of the hidden network $\mathcal{A}$ of the dimension $N_s*N_s$, are 
Gaussian random numbers generated before training, given the network size $N_s$, 
network link probability $d_r$, and spectral radius $\rho_r$. The reservoir network 
size $N_s$ is typically much larger than input dimension $D_i$ to ensure that it has 
sufficient capacity to learn the complex dynamics underlying the input signal. The 
output matrix $\mathcal{W}_{\rm out}$ has the dimension of $D_o * N_s$. During the 
training phase, the reservoir state $\mathbf{r}(t)$ is updated according to 
Eq.~\eqref{eq:reservoir} and is concatenated into a matrix $\mathcal{R}$ of the
dimension $N_s*T_l$, where $T_l$ is the total training length. With the 
corresponding input concatenated matrix $\mathcal{U}$, the output matrix 
$\mathcal{W}_{\rm out}$ can be obtained by Tikhonov 
regularization~\cite{bishop1995training} as
\begin{align} \label{eq:wout}
        \mathcal{W}_{\rm out}=\mathcal{U}\cdot \mathcal{R}^\intercal (\mathcal{R}\cdot \mathcal{R}^\intercal + \beta_r \mathcal{I})^{-1},
\end{align}
where $\beta_r$ is the regularization coefficient, $\mathcal{I}$ is the identity 
matrix of the dimension $N_s$. In the testing phase, the trained model takes the 
input $\mathbf{i}(t)$, updates the reservoir state $\mathbf{r}(t)$, and predicts 
the output $\mathbf{o}(t)$ via a linear combination of the reservoir state:
\begin{align} \label{eq:rc_out}
\bf{o}(t)=\mathcal{W}_{\rm out} \bf{r}(t).
\end{align}
As the ground truth data is no longer provided to the reservoir during the testing 
phase, the output from Eq.~\eqref{eq:rc_out} is fed back as the input at the next 
time step and used to update the hidden states. This process is repeated for any 
long time until the desired length of dynamical prediction is reached.

\section{Hyperparameter optimization}  \label{appendix_c}

Hyperparameter values are essential for machine-learning methods, which can have
a significant effect on the performance. We optimize the hyperparameters of 
transformer and reservoir computing through random search and Bayesian optimization, 
respectively.

\subsection{Transformer hyperparameter optimization}

\begin{figure*} [ht!]
\centering
\includegraphics[width=\linewidth]{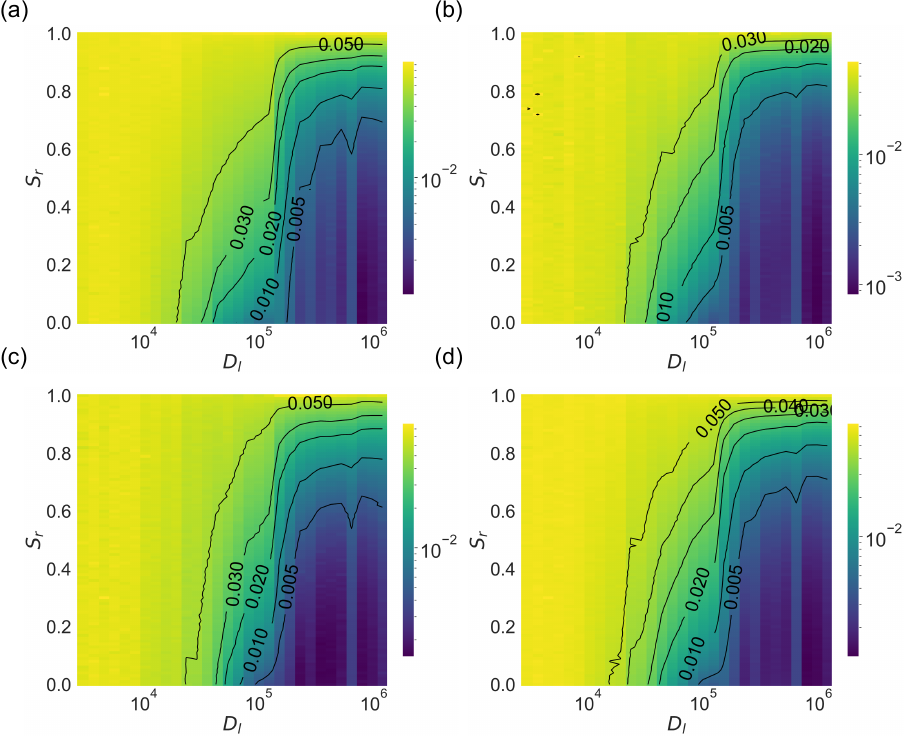} 
\caption{Effects of the training length $D_l$ on dynamics recovery performance, 
for different sparsity $S_r$. (a-c) Color-coded ensemble-averaged MSE values in 
the plane of ($D_l,S_r$) for the food-chain, Lorenz, and Lotka-Volterra systems, 
respectively. (d) Averaged performance over the three systems. Each value of the 
averaged MSE is obtained using 50 statistical realizations. Increasing the training 
length can often dramatically improve the prediction performance.}
\label{fig:net_size}
\end{figure*}

\begin{figure*} [ht!]
\centering
\includegraphics[width=\linewidth]{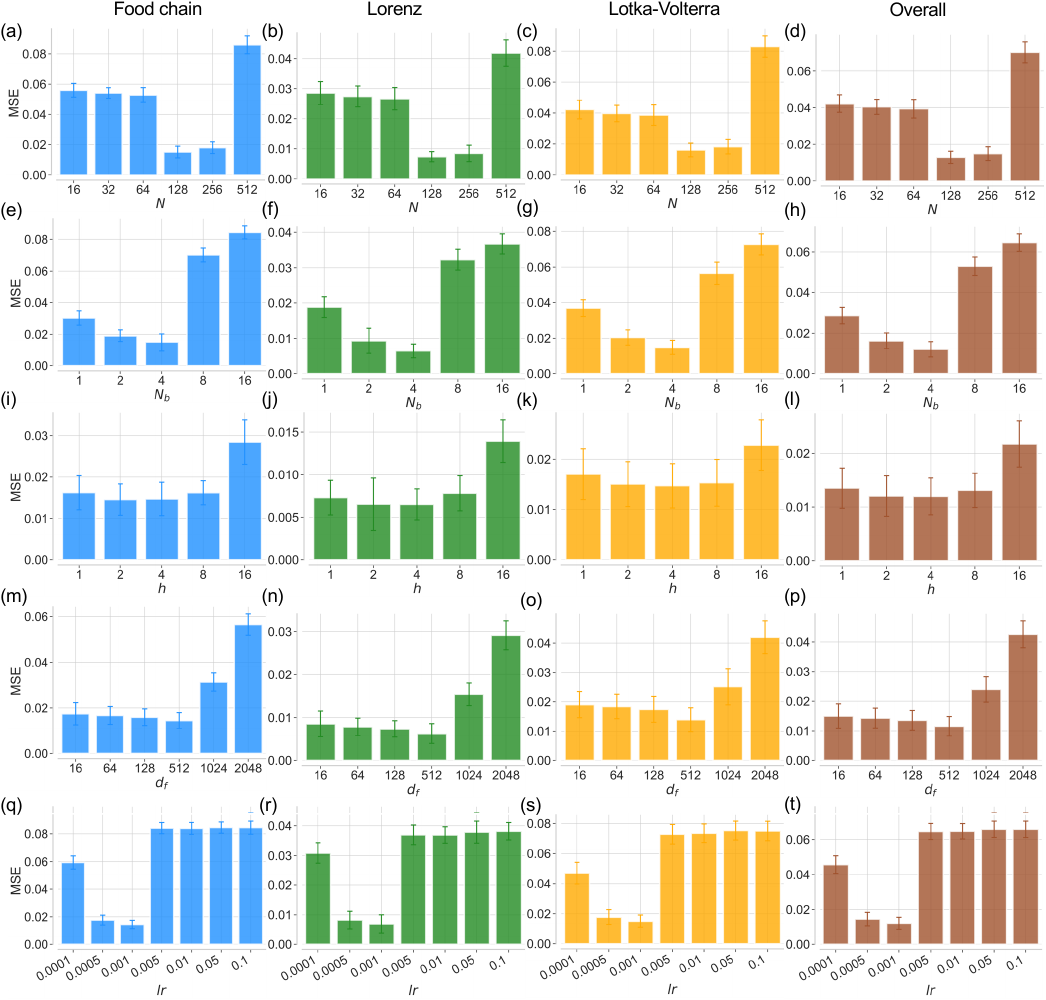} 
\caption{Effects of representative hyperparameters on performance. Columns 1-4 
are the results for the food chain, Lorenz, Lotka-Volterra systems, and the 
averaged performance of these systems, respectively. Rows 1-5 present the 
performance with respect to the following hyperparameters: input embedding 
dimension $N$, number $N_b$ of transformer blocks, transformer heads $h$, 
feedforward neurons $d_f$, and learning rate $lr$, respectively. To reduce 
the statistical fluctuations, 50 transformer models are trained to obtain the 
averaged MSE.}
\label{fig:hyper_transformer}
\end{figure*}

Random search is a simple yet effective method for hyperparameter optimization. 
Unlike grid search where hyperparameters are tested exhaustively over each possible 
combination, random search allows sampling hyperparameter values in a larger space. 
Each time, it randomly selects a combination of the hyperparameters from the search 
space, evaluates the performance on the validation dataset, and chooses the best 
performance combination. Due to the large number of hyperparameters the transformer 
contains, random search is practically effective in terms of the trade-off between 
the computational cost and performance. 

We optimize eleven hyperparameters for transformer through random search, with their
values listed in Table~\ref{tab:hyper_t}. To gain an intuitive understanding of 
how these hyperparameter values impact the performance, we test the target systems 
as they are not used in training and therefore are not used during the hyperparameter 
optimization process. We compare the prediction performance between the two cases
where the hyperparameters are optimized and not. Two chaotic Sprott systems, 
Sprott0 and Sprott1, are used to find optimal hyperparameters. As transformer is 
typically ``data-greedy,'' we not only prepare multiple systems, but for each 
system, we also provide a sufficient amount of data for training. To demonstrate
statistically how much data is required from each training system, we use varying 
training data length $D_l$ and evaluate the model performance on target systems. 
Representative results are shown in Fig.~\ref{fig:net_size}. It can be seen that 
the MSEs are large when $D_l$ is small, but after reaching a threshold about 
$D_l=10^5$, the performance improves dramatically. Notably, the sparsity affects 
the model performance as well: if it is large, increasing $D_l$ hardly yields 
any improvement. This is reasonable, as merely increasing the training data for 
a specific system does not enhance the transformer's ability to generalize. In fact, 
it may even hinder the capacity of the model to learn new dynamics due to overfitting 
on a specific system. 

To demonstrate the effectiveness of random search in finding optimal or near-optimal 
hyperparameter sets, we present the results on the effects of typical hyperparameters 
on the performance, as depicted in Fig.~\ref{fig:hyper_transformer}, where all 
hyperparameters are fixed except one. We train the transformer multiple times to 
evaluate the effect of varying this hyperparameter on the averaged MSE across 
trained systems. The results show that in most cases, the hyperparameters determined 
by random search are indeed optimal or near-optimal for training the transformer.

\subsection{Reservoir computing hyperparameter optimization}

\begin{figure} [ht!]
\centering
\includegraphics[width=\linewidth]{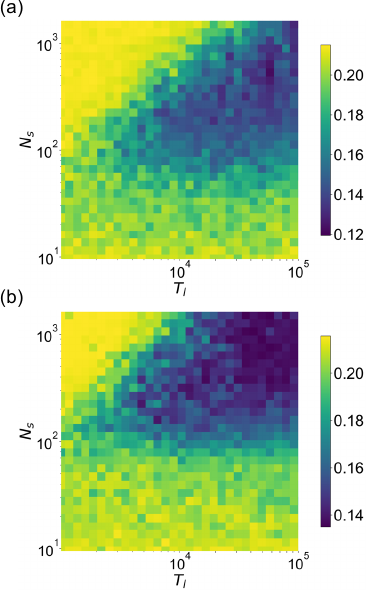} 
\caption{Effects of reservoir-computing training length and network size on 
performance. (a,b) Ensemble-averaged MSE obtained from 50 realizations versus 
changes in the training length $T_l$ and network size $N_s$ for the chaotic 
Lorenz and Lotka-Volterra systems, respectively. Appropriate parameter ranges 
lead to satisfactory performance.}
\label{fig:hyper_rc}
\end{figure}

A key issue of reservoir computing is that its performance often depends 
sensitively on the hyperparameters. We apply Bayesian optimization from 
Python (\textit{bayesian-optimization}) to find the optimal values of the 
following hyperparameters: leakage parameter $\alpha_r$, regularization 
coefficient $\beta_r$, input matrix scaling factor $\gamma_r$, spectral radius 
$\rho_r$, link probability $d_r$ determining the network matrix $\mathcal{A}$, and 
the amplitude $\sigma_r$ of the noise added to the input signal. The optimized 
hyperparameters are listed in Tab.~\ref{tab:hyper_rc}. We evaluate the performance 
of the reservoir-computing based long-term ``climate'' or attractor reconstruction 
with respect to varying the training length $T_l$ and network size $N_s$. 
Figures~\ref{fig:hyper_rc}(a) and \ref{fig:hyper_rc}(b) show the performance 
versus the two hyperparameters for the Lorenz and Lotka-Volterra systems, 
respectively. It can be seen that a combination of larger network and longer 
training length will lead to better reconstruction performance, while increasing 
the network size for small $T_l$ will degrade the performance. 

\begin{table}[ht!]
\caption{\label{tab:hyper_rc}
Optimal hyperparameter values of reservoir computing for target systems}
\begin{ruledtabular}
\begin{tabular}{ccccccc}
System & $\alpha_r$ & $\beta_r$ & $\gamma_r$ & $\rho_r$ & $d_r$ & $\sigma_r$\\
\hline \\
Food chain & 0.36 & $10^{-1.25}$ & 1.16 & 1.29 & 0.41 & $10^{-4.70}$\\
Lorenz  & 0.30 & $10^{-5.15}$ & 1.82 & 1.30 & 0.68 & $10^{-2.04}$\\
Lotka-Volterra & 0.29 & $10^{-6.62}$ & 0.19 & 1.72 & 0.02 & $10^{-2.73}$\\
\end{tabular}
\end{ruledtabular}
\end{table}

\section{Further demonstration of dynamics reconstruction}  \label{appendix_d}

\subsection{Additional examples of dynamics reconstruction}

\begin{figure*} [ht!]
\centering
\includegraphics[width=\linewidth]{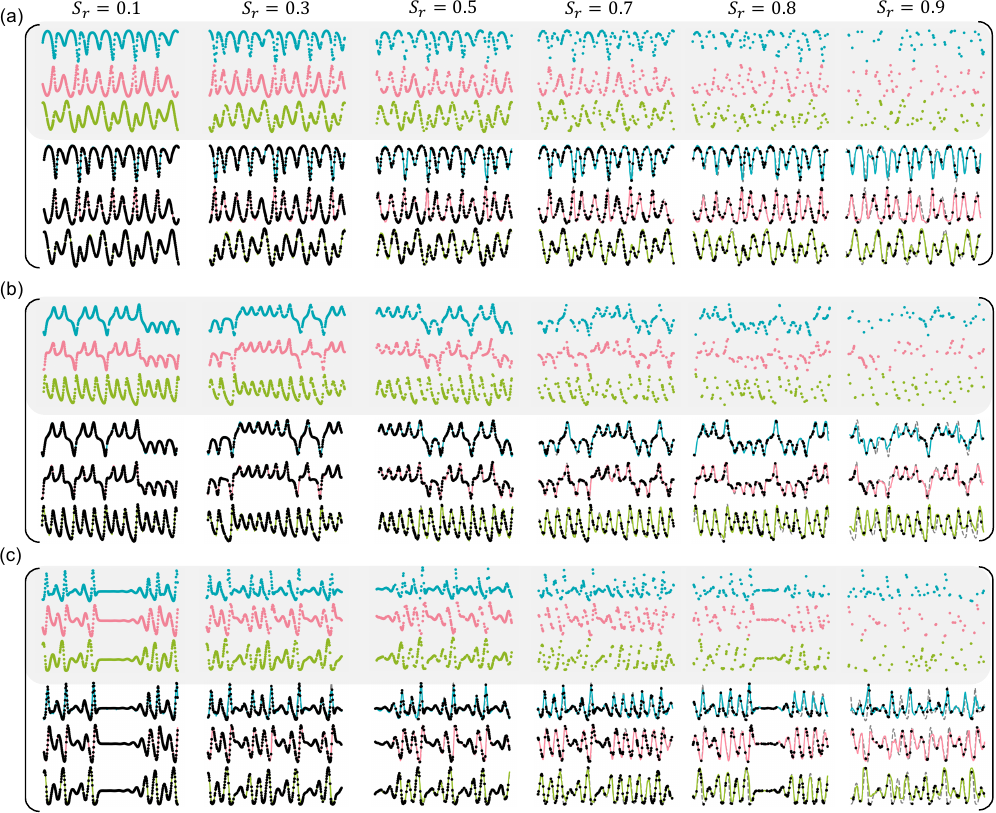} 
\caption{Additional examples of dynamics reconstruction by transformer for 
different levels of data sparsity. (a-c) Reconstructed time series for the 
food chain, Lorenz, and Lotka-Volterra systems, respectively, where the shaded 
areas represent observable points and unshaded areas show the corresponding 
reconstruction results. The sequence length is fixed at $L_s=1200$ (around 25 cycles of oscillations), but only 
the first 500 data points are displayed for clarity of presentation.}
\label{fig:systems_mr}
\end{figure*}

\begin{figure*} [ht!]
\centering
\includegraphics[width=\linewidth]{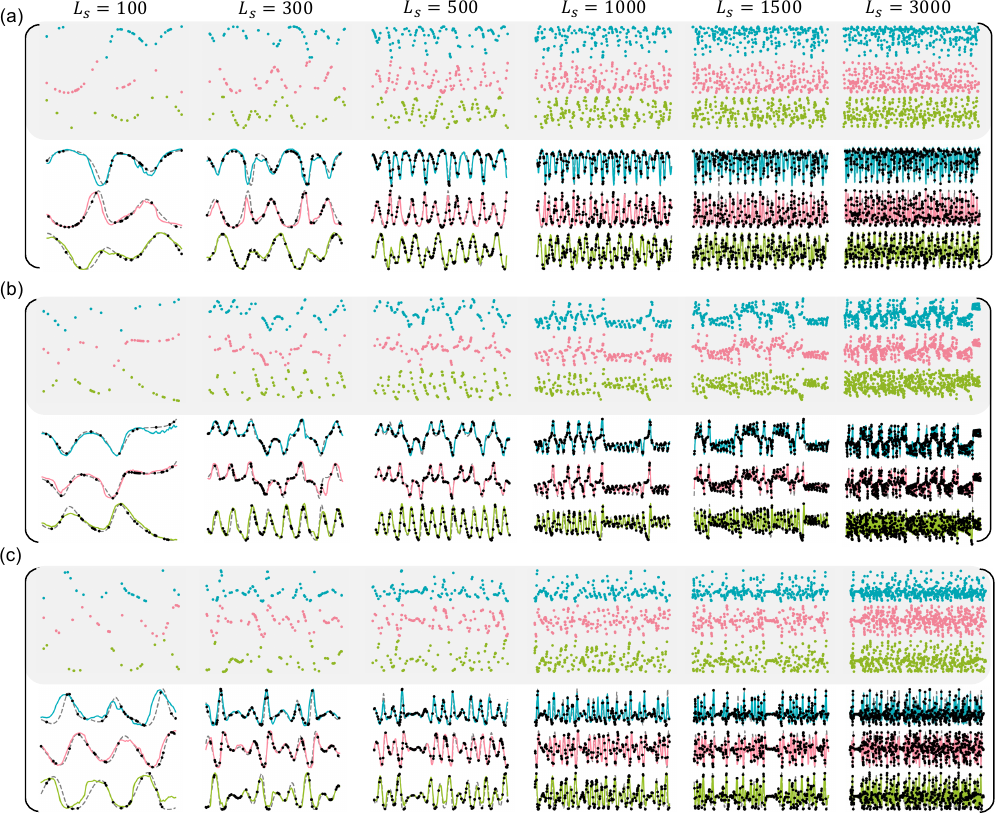} 
\caption{Additional examples of dynamics reconstruction with respect to varying 
input sequence length $L_s$. (a-c) Reconstructed time series for the food chain, 
Lorenz, and Lotka-Volterra systems, respectively, where the shaded areas represent 
the observable points and unshaded areas show the corresponding reconstruction. 
The sparsity value is fixed at $S_r=0.8$.}
\label{fig:systems_ls}
\end{figure*}

In the main text, we presented examples of the dynamics reconstruction of the 
food chain system. To demonstrate that the transformer-based framework performs
for varying sparsity $S_r$ and sequence length $L_s$, we present additional 
examples for the food chain system, as well as the Lorenz and Lotka-Volterra 
systems, as shown in Figs.~\ref{fig:systems_mr} and \ref{fig:systems_ls}, with
results comparable to those in the main text: as $L_s$ increases and $S_r$ 
decreases, the performance of the model on target systems continuously improves.

\subsection{Performance of system dynamics recovery}

\begin{figure*} [ht!]
\centering
\includegraphics[width=\linewidth]{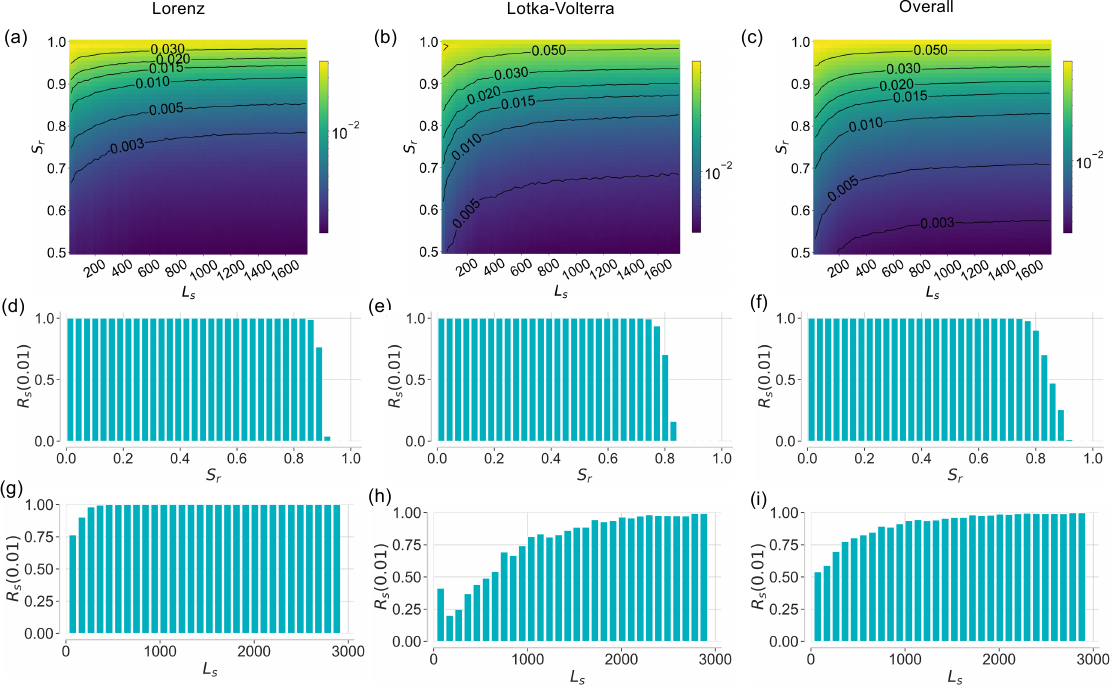} 
\caption{Dynamics reconstruction performance. The three columns are the results 
for the Lorenz, and Lotka-Volterra systems, and the averaged performance over the 
three target systems, respectively. The three rows from top to down present the 
color-coded averaged MSE in the parameter plane ($L_s,S_r$), the recovery 
stability $R_s$ versus $S_r$ and $L_s$ for MSE threshold 0.01, respectively. 
Reconstruction stability and averaged MSE are calculated from 400,000 data points
(corresponding to around 8,000 cycles of oscillations) in each case.}
\label{fig:performance}
\end{figure*}

In addition to the food chain system in the main text, we further evaluate the 
performance of the other two target systems. Figure~\ref{fig:performance} 
illustrates the reconstruction performance for the Lorenz and Lotka-Volterra 
systems, as well as the overall performance across all three target systems, 
as characterized by MSE [see Eq.~\eqref{eq:mse}] and recovery stability 
[Eq.~\eqref{eq:recovery_stability}]. For reconstruction stability analyses, 
$L_s$ is fixed at 1,200 (around 25 cycles of oscillations) when varying $S_r$, and $S_r$ is set to 0.8 when 
varying $L_s$. For the Lotka-Volterra system, we set $S_r$ to 0.75 to ensure 
reconstruction stability reaching one given sufficient $L_s$. These results 
further demonstrate the ability of the transformer to reconstruct the dynamics 
of new systems from sparse observational data.

\subsection{Performance of long-term ``climate'' reconstruction}

\begin{figure*} [ht!]
\centering
\includegraphics[width=0.8\linewidth]{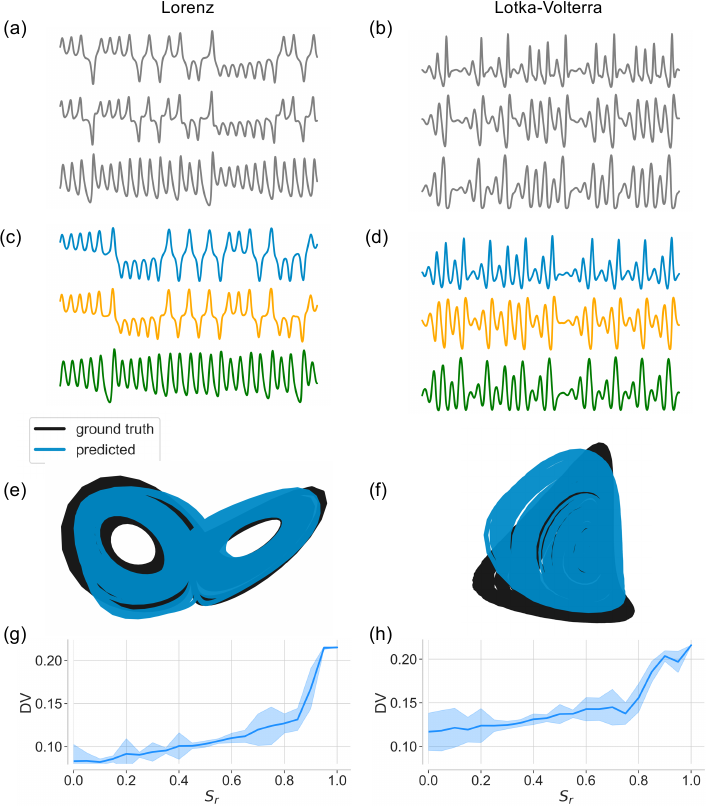} 
\caption{Attractor reconstruction by reservoir computing. The first and second 
columns are for the Lorenz and Lotka-Volterra systems, respectively. (a,b) Segments 
of ground truth time series, (c,d) short-term predictions by reservoir computing, 
(e,f) reconstructed long-term attractors as compared with the true attractors, and
(g, f) ensemble-averaged DV versus sparsity $S_r$, obtained from 50 independent
realizations.}
\label{fig:rc_examples}
\end{figure*}

Reservoir computing utilizes and trains on the reconstructed time series from 
the transformer and is able to generate arbitrarily long time series with the
same statistical properties as those of the original system. Here we present 
results from long-term reconstruction of the Lorenz and Lotka-Volterra systems, 
where the time series reconstructed from the transformer are for $S_r=0.8$ and 
0.75, respectively. Figures~\ref{fig:rc_examples}(a) and \ref{fig:rc_examples}(b)
show segments of the ground truth for the Lorenz and Lotka-Volterra systems, 
respectively. Figures~\ref{fig:rc_examples}(c) and \ref{fig:rc_examples}(d) 
show the time series generated from reservoir computing for the respective 
systems. The reconstructed (predicted) and ground truth chaotic attractors 
of the two systems are depicted in Figs.~\ref{fig:rc_examples}(e) and 
\ref{fig:rc_examples}(f), respectively. It can be seen that the reservoir
computer can capture the ``climate'' through the transformer output, providing
indirect confirmation that the transformer has successfully reconstructed the 
dynamics. Figures~\ref{fig:rc_examples}(g) and \ref{fig:rc_examples}(h) 
show the DV versus varying $S_r$. In general, as the observations become more 
sparse, the attractor reconstruction performance as measured by DV gradually 
degrades. For $S_r>0.8$, the performance deteriorates rapidly.

\section{Robustness test} \label{appendix_e}

\begin{figure*} [ht!]
\centering
\includegraphics[width=\linewidth]{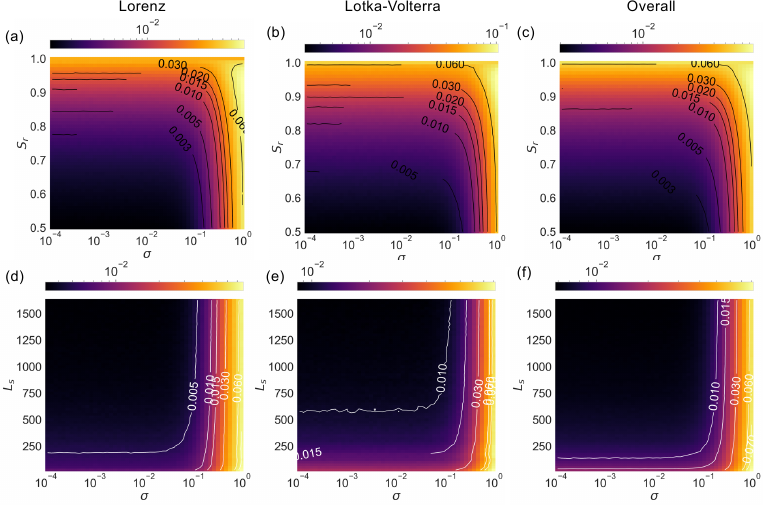} 
\caption{Robustness against multiplicative noise. Three columns represent results 
for the Lorenz, and Lotka-Volterra systems, and the averaged performance over the 
three target systems, respectively. The two rows show ensemble-averaged, color-coded 
MSE from 50 independent realizations, over the parameter planes ($\sigma, S_r$) 
and ($\sigma, L_s$), respectively. The results indicate that the transformer-based 
dynamics-reconstruction framework is robust against multiplicative noise.}
\label{fig:noise}
\end{figure*}

\begin{figure*} [ht!]
\centering
\includegraphics[width=0.6\linewidth]{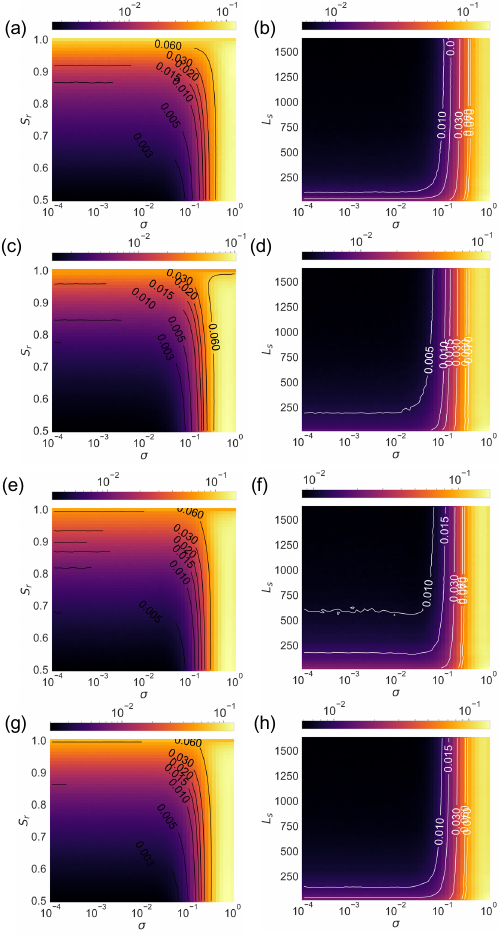} 
\caption{Robustness against additive noise. The four rows represent results for the 
food chain, Lorenz, Lotka-Volterra systems, and the overall performance, 
respectively. The two columns show ensemble-averaged, color-coded MSE from 50 
independent realizations, over the parameter planes ($\sigma, S_r$) and 
($\sigma, L_s$), respectively. The transformer-based dynamics reconstruction 
framework is robust against additive noise.}
\label{fig:noise_add}
\end{figure*}

We present the results of robustness test against multiplicative and additive 
noise, for the three target systems, as shown in Figs.~\ref{fig:noise} and 
\ref{fig:noise_add}. It can be seen that the transformer based dynamics 
reconstruction framework is robust against multiplicative and additive 
noise of amplitude below $10^{-1}$ and $10^{-1.5}$, respectively. The sequence
length is $L_s = 1200$ (around 25 cycles of oscillations) for varying $S_r$ and noise level $\sigma$, while 
$S_r = 0.8$ for varying $L_s$.

\section{A counter example} \label{appendix_f}

\begin{figure*} [ht!]
\centering
\includegraphics[width=\linewidth]{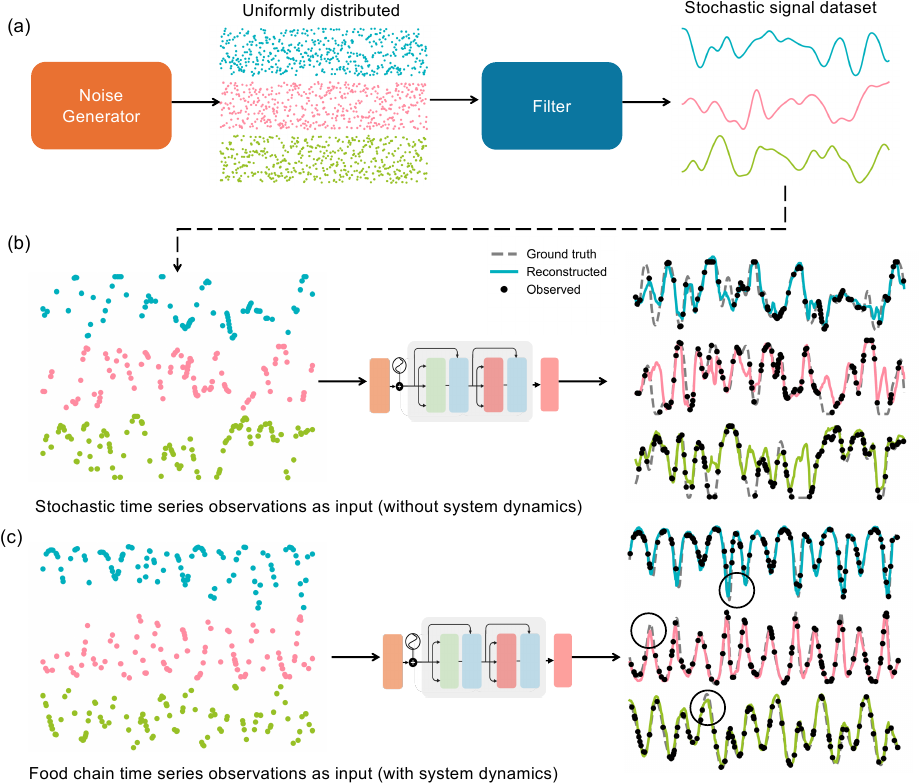} 
\caption{Illustration of stochastic signal recovery. (a) Generation of the 
stochastic signal dataset. (b,c) Examples of reconstructing time series without and with underlying system dynamics by using a well-trained transformer 
for $L_s=1200$ (around 25 cycles of oscillations) and $S_r=0.8$.} 
\label{fig:stochastic_illustration}
\end{figure*}

\begin{figure*} [ht!]
\centering
\includegraphics[width=\linewidth]{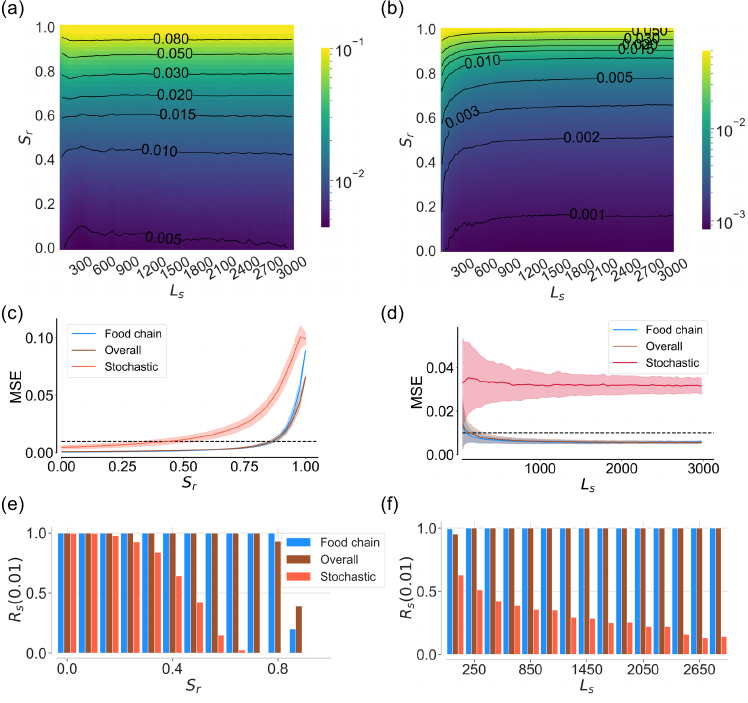} 
\caption{Performance of transformer on stochastic signals and target systems. 
(a,b) Ensemble-averaged color-coded MSE in the parameter plane ($L_s, S_r$) for 
the stochastic and target dynamical systems, respectively, (c) MSE versus $S_r$ 
for $L_s = 1200$ (around 25 cycles), (d) MSE versus $L_s$ for $S_r=0.8$, (e) reconstruction stability 
$R_s(\cdot)$ versus $S_r$ for $L_s=1200$, (f) reconstruction stability $R_s(\cdot)$ 
versus $L_s$ for $S_r=0.5$. The reconstruction stability and the averaged MSE are 
calculated from 400,000 data points (corresponding to around 8,000 cycles) in each case.}
\label{fig:stochastic_performance}
\end{figure*}

For testing, three target systems are presented as examples of reconstructing
the system dynamics, assuming that the transformer has never been exposed to these
dynamics. This indicates that the transformer trained using data from a large
number of synthetic dynamical systems already possesses the ability to find the 
dynamics behind the sparse data from the target system. To verify that this is 
indeed the case, we study two scenarios for comparison: (1) the transformer has
such a ``dynamics-adaptable'' ability and (2) it does not have the ability.
For the first scenario, we use stochastic signals for testing. Specifically, we 
generate uniformly distributed noise independently for the three dimensions. A 
Gaussian filter with the standard deviation $\sigma_g=12$ of the Gaussian kernel 
is applied to smooth the generated noisy data. The resulting stochastic signals 
are normalized and collected as the testing dataset, as shown in 
Fig.~\ref{fig:stochastic_illustration}(a). We then take a well-trained transformer 
adapted for other chaotic systems and test it with the stochastic signal, as shown 
in Fig.~\ref{fig:stochastic_illustration}(b). For comparison, performance with 
the food-chain system is shown in Fig.~\ref{fig:stochastic_illustration}(c).
For the same sparsity $S_r$ and sequence length $L_s$, the transformer successfully 
reconstruct the dynamics of the food chain system but completely fails to reconstruct
the stochastic signal. The general conclusion is that, the sparse data presented
to the transformer need to be associated with some deterministic, nonlinear dynamical
process to achieve successful reconstruction. 

We further study the performance of the transformer using the stochastic signals 
and the dynamical systems, as shown Fig.~\ref{fig:stochastic_performance}. It can 
be seen that, regardless of the values of $S_r$ and $L_s$, the performance on 
the target dynamical systems consistently surpasses that on stochastic signals. 
A performance comparison for $S_r=0.5$ is shown in 
Fig.~\ref{fig:stochastic_performance}(f). Even with this low sparsity, the
transformer is unable to reconstruct the stochastic signal. 

\section{Reconstructing dynamics of diverse chaotic systems} \label{appendix_g}

\begin{figure*} [ht!]
\centering
\includegraphics[width=\linewidth]{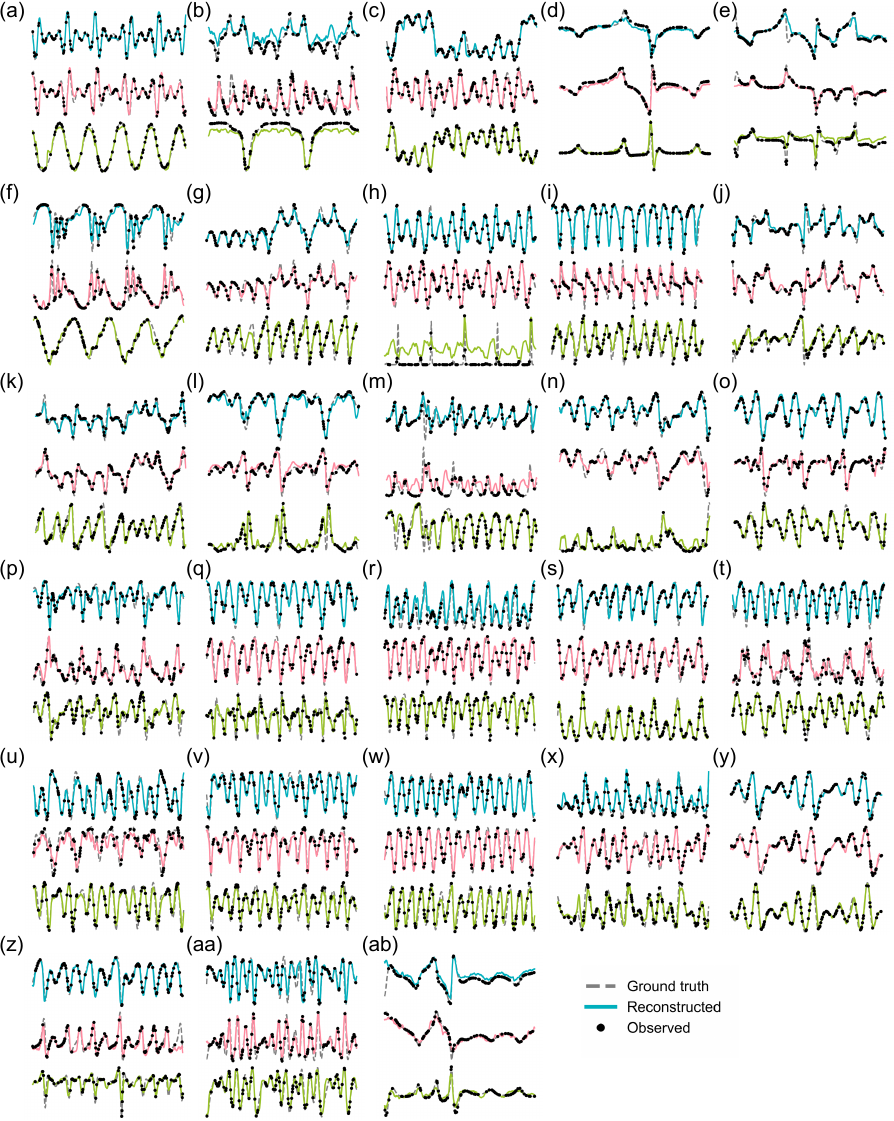} 
\caption{Dynamics reconstruction of 28 different chaotic systems: (a) Aizawa, 
(b) Bouali, (c) Chua, (d) Dadras, (e) Four wing, (f) Hastings-Powell, (g) Rikitake, 
(h) Rossler, (i-aa) Sprott systems. (ab) Wang system, for $S_r=0.8$ and $L_s=1200$ (around 25 cycles).}  
\label{fig:all_systems}
\end{figure*}

\begin{figure*} [ht!]
\centering
\includegraphics[width=0.68\linewidth]{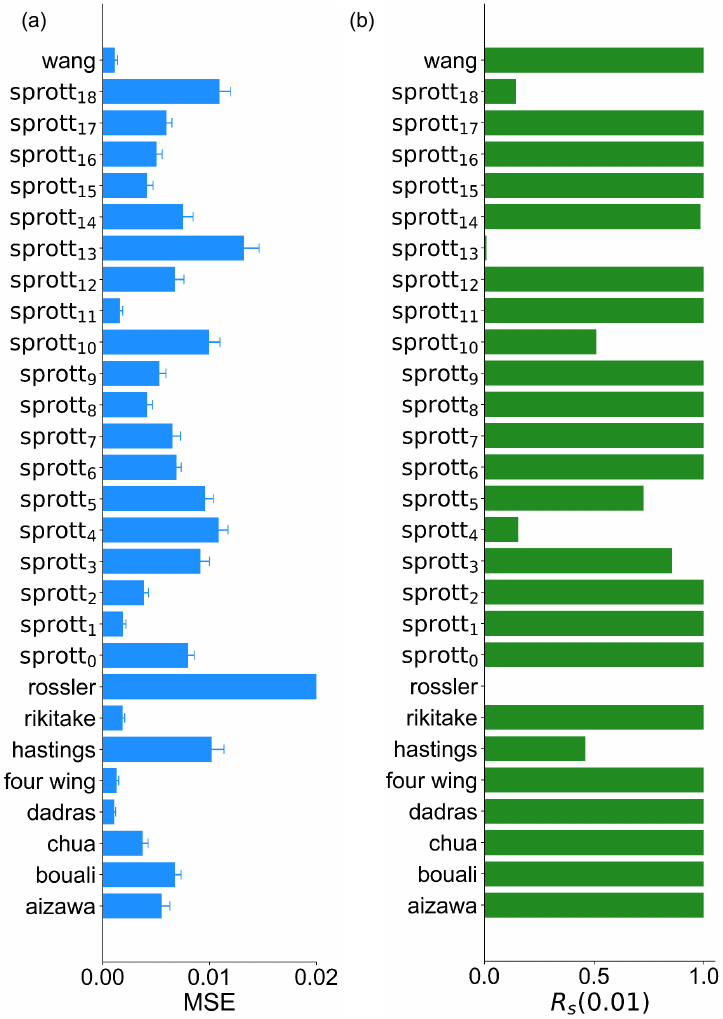} 
\caption{Performance of dynamics reconstruction on the 28 chaotic systems. 
(a) Averages MSEs calculated from 400,000 data points (corresponding to about 
8,000 cycles of oscillations) in each case. (b) Reconstruction stability 
$R_s(\cdot)$ for $S_r=0.8$ and $L_s=1200$ (about 25 cycles).}
\label{fig:all_systems_performance}
\end{figure*}

In the main text, we have used 28 chaotic systems for training the transformer 
and three specific systems for testing. Altogether, 31 chaotic systems have been
used. Here we demonstrate that the transformer can also reconstruct arbitrary 
systems. In particular, each time we choose several systems as the targets for
testing, which are excluded from the list of training systems. The remaining 
chaotic systems are used for training. After training, sparse data from the 
chosen target system are presented to the transformer for reconstruction. Repeating 
this process, each and every system in the list of 31 systems can be tested as
the target. Representative reconstruction results are shown in 
Fig.~\ref{fig:all_systems}, where four systems are chosen as the targets each time.
For most systems, the transformer demonstrates can faithfully reconstruct the
target dynamics for $S_r = 0.8$ and $L_s = 1200$. However, for certain target 
systems, there is reduced performance. For instance, the construction of the third 
dimension of the chaotic R\"{o}ssler system is unsatisfactory, in spite of the 
excellent reconstruction of the first and second dimensions, as shown in 
Fig.~\ref{fig:all_systems}(h). The reason is that the third dynamical variable 
of the R\"{o}ssler system exhibits a kind of impulsive behavior.
Figure~\ref{fig:all_systems_performance} shows the performance of dynamics 
reconstruction of the 28 different chaotic systems in terms of MSE (a) and 
reconstruction stability $R_s(\cdot)$ (b) for $S_r = 0.8$. The overall MSE is low
the reconstruction is highly stable across all the 28 systems, demonstrating the
power and generalizability of the transformer based reconstruction framework.

\section{Reconstructing dynamics with long short-term memory} \label{appendix_h}

\begin{figure*} [ht!]
\centering
\includegraphics[width=\linewidth]{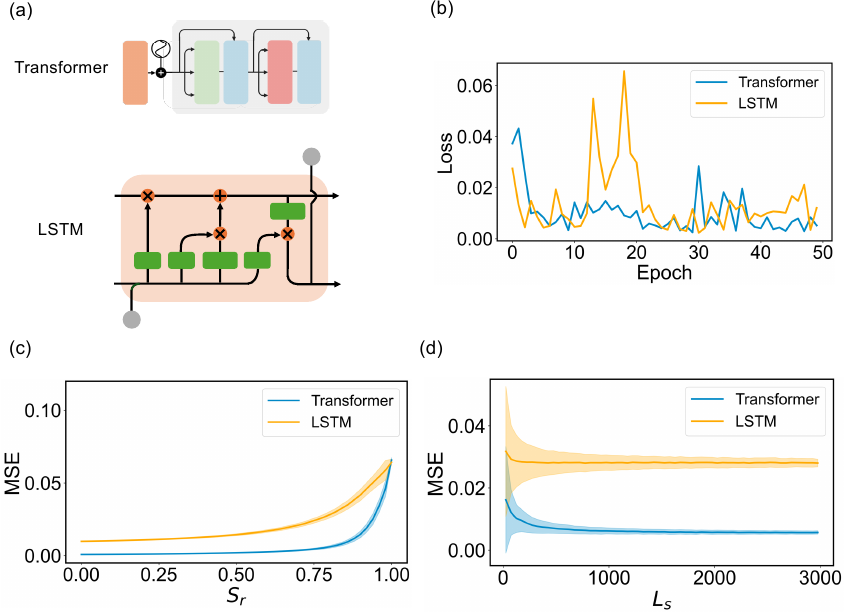} 
\caption{Comparison of dynamics reconstruction performance between transformer 
and LSTM. (a) Basic structures of the two machine learning frameworks. (b) Training 
loss of transformer and LSTM. (c) MSE versus $S_r$ for transformer and LSTM for
$L_s=1200$ (around 25 cycles). (d) MSE versus $L_S$ for $S_r=0.8$. Averaged MSEs are calculated 
from 400,000 data points (corresponding to around 8,000 cycles of oscillations) in each case, for the food chain, Lorenz, and 
Lotka-Volterra systems.}
\label{fig:compare_lstm}
\end{figure*}

Long short-term memory (LSTM) is a type of recurrent neural networks designed to 
capture the long-term dependencies in sequential data~\cite{hochreiter1997long}. 
LSTM has demonstrated remarkable success in time series forecasting, natural 
language processing, and other tasks, due to its unique cell structure comprising 
input, forget, and output gates, which regulate the flow of information through 
the network. The relevant information is retained over extended time steps in 
LSTM networks, helping them capture temporal dependencies. The basic structure of
LSTM is shown in Fig.\ref{fig:compare_lstm}(a).

We investigate whether LSTM can be an appropriate substitute for the transformer
for dynamics reconstruction from sparse data. To ensure a fair comparison, we 
keep the hyperparameters as consistent as possible. For example, the hidden 
network size is set to 512, and the number of layers is set to 4. We train 
both the LSTM and transformer using the same setup for each training instance, 
with randomly chosen sequence lengths $L_s$ and sparsity $S_r$, using a batch 
size of 16 and 50 epochs. The training loss versus the epoch is depicted in 
Fig.\ref{fig:compare_lstm}(b), indicating that, after the training, the two 
machine-learning frameworks exhibit similar performance on these training systems. 
However, for testing, the 
transformer yields much lower MSE values, as shown in Figs.~\ref{fig:compare_lstm}(c) 
and \ref{fig:compare_lstm}(d). These results suggest that, for dynamics 
reconstruction from sparse data of unseen dynamical systems, transformer is 
more effective than traditional recurrent neural networks.

\clearpage


%
\end{document}